\documentclass[11pt]{article}
\usepackage{amsmath}
\usepackage{amsfonts}
\usepackage[pdftex]{graphicx}
\usepackage{float}

\newcommand{\qed}{\nobreak \ifvmode \relax \else
      \ifdim\lastskip<1.5em \hskip-\lastskip
      \hskip1.5em plus0em minus0.5em \fi \nobreak
      \vrule height0.75em width0.5em depth0.25em\fi}

\begin{document}

\title{The Information Theoretically Efficient Model (ITEM): A model for computerized analysis of large datasets}
\author{Tyler Ward\\
  \texttt{ward.tyler@gmail.com}}
\date{\today}
\maketitle

\pagebreak

\tableofcontents

\pagebreak

\begin{abstract}

This document discusses the Information Theoretically Efficient Model (ITEM), a computerized system to generate an information theoretically efficient multinomial logistic regression from a general dataset. More specifically, this model is designed to succeed even where the logit transform of the dependent variable is not necessarily linear in the independent variables. This research shows that for large datasets, the resulting models can be produced on modern computers in a tractable amount of time. These models are also resistant to overfitting, and as such they tend to produce interpretable models with only a limited number of features, all of which are designed to be well behaved. 

\end{abstract}

\section{Introduction}

\paragraph{}

A good overview of the state of the art can be found in \cite{NELDER80}. That paper introduces the general problem ITEM was designed to solve, and very briefly surveys the various methods available in the literature that can solve this problem. 

\subsection{Statement of the Problem}

\paragraph{}

Consider the usual regression situation: we have data $ (\vec x_i, \vec y_i) $ for $ i = 1, 2, ..., N $. Here $ \vec x_i = (x_{i,1}, x_{i,2}, ...., x_{i,K})^T $ and $ \vec y_i = (y_{i,1}, y_{i,2}, ... y_{i,W})^T $ are the regressor and response variables. Suppose also that $ x_{i,k} \in \mathbb{R} $ and $ \vec y_i $ is a probability vector. In shorthand, $ X = \{\vec x_{i}\} $ and $ Y = \{ \vec y_{i} \} $. 

\paragraph{}

A model is some function $ A_{XY}(x_w) $ whose output is a prediction for $ y_w $. In most common situations, the model $ A_{XY} $ must be deduced from the data $ X = \{\vec x_1, \vec x_2, ....., \vec x_N \} $ and $ Y = \{ \vec y_1, \vec y_2, ..., \vec y_N \} $, this process is referred to as fitting. Then the model would be used to make predictions for $ \vec x_w $ with $ w > N $, i.e. predict $ \vec y $ for the data points where we only know $ \vec x $, this is referred to as projection or running. ITEM is one such model, but several others will also be discussed. 

\paragraph{}

In addition to the requirements above, a two more factors are added from an engineering standpoint:

\begin{itemize}
\item A large amount of data is available (at least 100,000 observations)
\item The data has more than a few dimensions ($ \vec x_i $ has at least 5 dimensions)
\end{itemize}

For one dimensional data, various solutions are available, but generally, people will simply look at the results manually and make any adjustments that are needed. Beyond a few dimensions, this no longer works efficiently. It uses too much labor, and the projections on to the individual dimensions become less and less tractable. Building up a model from a set of $ N $ regressors requires roughly $ N^{2} $ operations, as for each regressor added, all other regressors need to be checked. With a moderate number of regressors (dozens to hundreds) this is too time consuming for people, but could be done by a computer. The primary goal of ITEM is to perform this multidimensional analysis and fitting automatically. Since human guidance is limited, it is important for the model to be able to resist overfitting while simultaneously using the data efficiently to produce a model as accurate as possible. The dataset must be large, because it will generally require a lot of data to support good predictions across many dimensions. 

\subsubsection{The Analogy}

\paragraph{}

Before proceeding further, this entire paper is devoted to producing good models. It is worth taking a brief detour to discuss what is meant by a good model. For this purpose, it is helpful to use a car analogy. For a person buying a car, there are several statistics that might be important. 

\begin{enumerate}
\item acceleration
\item top speed
\item fuel efficiency
\item material efficiency
\item automated construction
\end{enumerate}

\paragraph{}

Each of these factors helps drive the appeal (and cost) of a given car. For instance, a car could be made out of carbon fiber and titanium, thus getting better speed and efficiency, but by using these extremely expensive materials, the cost may be prohibitive. Similarly, a car could be constructed with extremely advanced manufacturing techniques, but if it cannot be mass produced, then again the car would be too expensive for typical use. Among these five characteristics, often improving one will make others worse. 

\paragraph{}

This tradeoff will define a sort of efficient frontier, a 5 dimensional surface enclosing some volume. Each point on this frontier represents some tradeoff among the 5 qualities, but there is no point reachable with this technology that is better in all 5 categories. One definition of the quality of automotive technology is to simply say that better technology allows for a larger frontier, one that completely encloses the frontier for an inferior set of technology. A model has a similar set of factors that together help to determine its quality. 

\begin{enumerate}
\item fitting computational efficiency
\item running computational efficiency
\item data efficiency
\item parameter efficiency
\item automated construction
\end{enumerate}

\paragraph{}

A good model would be computationally inexpensive to fit and to run. Just like in the case of a car, these are nearly the same requirement. A car with good acceleration will typically have a high top speed, similarly a model that is efficient to fit will typically (though not always) also be efficient to run. These two factors will be grouped together as computational efficiency. 

\paragraph{}

Data efficiency and parameter efficiency are similar concepts as well, so they will be grouped together as information theoretic efficiency. Data efficiency means essentially that statistically significant features of the data will be included in the model. We require that a feature will be picked up by the model without requiring excessive amounts of data, it will be added as soon as enough data is present for it to be significant. The concept of parameter efficiency means that the model will not pick up extra features beyond those indicated by the data. Essentially, this is a requirement that the model use as few parameters as possible, while simultaneously fitting the data as well as possible. Obviously there is some trade off between these two features, but together they define a part of the efficient frontier for models. We would like the model to be on or near this frontier, and for this frontier to itself pass close to the true distribution of the data. 

\paragraph{}

The last requirement is simply that the model be constructed automatically, with limited human involvement. The whole purpose of this exercise is to produce a good model without expending a large amount of human effort to do so. 

\paragraph{}

It is not possible for a model to excel in all five of these areas. Some models are better in one, or another, and for most distributions of data there is a definite limit to how high the information theoretic efficiency of a model can be, just as a car's fuel efficiency runs into thermodynamic limits. The goal of the ITEM system is to construct models that are very good (though by no means ideal) in all 5 of these areas.

\subsubsection{The Example Problem}
\label{example}

\paragraph{}

For the purposes of this paper, a residential mortgage model will be considered. In the United States, a residential mortgage is a loan made by a lender (typically a bank) to a borrower (typically an individual) for the purpose of buying residential real estate, typically a house, but sometimes a condo or other living space. When making such a loan, it is important for the lender to be able to project the probable future behavior of the borrower. In order to do that, the lender can examine a large dataset of historical loan behavior, and attempt to build a model based on this historical record. 

\paragraph{}

One such dataset is the Freddie Mac loan level dataset \cite{FREDDIE}, which will be used to fit this example model. This data set has approximately 600 million observations, so it is large enough to show an example of a real model. Other similar data sets are also available, most of which contain one to several billion observations. Since this is real data, projections based on it have actual economic value. For instance, projections of future behavior determine the interest rate offered to the borrower, and help the lender determine how to hedge efficiently.   

\paragraph{}

The dataset is a collection of loan-month observations. By this, it is meant that each loan in the dataset has one observation per month. This observation is little more than the knowledge of whether or not the borrower made his payment that month. These observations will be considered independently, ignoring the fact that some of them are related to each other through the underlying loan. The logic here is that if the model is accurate and the dataset is comprehensive, then the loan-months will be independent conditional on the data presented. Another way of saying this is that the model residuals will be independent. This is an assumption that can be relaxed, but that is outside the scope of this paper.  

\paragraph{}

Each of these loan-months has several regressors, including FICO, several flags related to the type of mortgage, loan age, Debt-To-Income ratios, refinance incentive, and others. Each loan-month also has a status. This status is a combination of the number of payments the borrower is in arrears, as well as some additional information about potential foreclosure proceedings and similar items. For the purposes of this paper, only three statuses will be considered. 

\begin{itemize}
\item Status "C": The borrower is "current", and is not behind on his payments
\item Status "P": The borrower has repaid the entire loan, the loan no longer exists
\item Status "3": The borrower is one payment behind schedule ("30 days delinquent")
\end{itemize}

\paragraph{}

Now consider the status that the loan will have next month. If the loan-month is historical data, this value may be known, otherwise it it necessary to project it. The purpose of the example model is to predict next month's status for any loan-month in which we know this month's status. In this way, a Markov Chain can be built up, progressively projecting the status further and further into the future. 

\paragraph{}

In each month, the borrower transitions from his current delinquency state to a new state depending on the number of payments made. For a borrower who is Current (status "C"), he can make 1 payment (remaining "C"), 0 payments (to become "3"), or all the payments (to become "P"). Similarly, if a borrower is already in status "3", he could become more delinquent by missing additional payments and so on. A separate model can be fit for each loan status, modeling the transitions available to loan-months in that status, so for the purpose of this paper it is enough to consider only the status "C". 

\paragraph{}

Therefore, there are three transitions available, in shorthand written thus: 

\begin{itemize}
\item $ C \rightarrow C $ Going from Current to Current
\item $ C \rightarrow P $ Going from Current to Prepaid
\item $ C \rightarrow 3 $ Going from Current to 30 days delinquent
\end{itemize}

\paragraph{}

The transition $ C \rightarrow C $ is more common than the others by several orders of magnitude, but most of the economic behavior of these loans is related to the $ C \rightarrow P $ and $ C \rightarrow 3 $ transitions. 

\paragraph{}

Long before large datasets were available, mortgages were modeled using pool level approaches, which simply averaged a group of loans together and then applied the model. Some of these models are still around, however they are quickly falling out of favor. The essential issue with the pool level approach is that it is subject to numerous problems related to this averaging. For instance, there is a world of difference between two pools with 700 FICO, one where every loan has exactly a 700 and another where half the loans have 600 and the other half have 800. In many cases, the pool behavior is driven by just a handful of loans, and the information about this handful of loans is destroyed by the averaging process. 

\pagebreak

\subsubsection{The Example Data}

\begin{table}[H]
\caption {Example Mortgage Data}
\begin{center}
\begin{tabular}{|c|c|c|c|c|c|c|c|c|}

\hline
loanId & month & FICO & MTMLTV & age & incentive & isOwner & startStatus & endStatus \\
\hline
1 & 2014-01 & 750 & 75 & 14 & -0.3 & 1 & C & C \\
1 & 2014-02 & 750 & 75 & 15 & -0.35 & 1 & C & P \\
2 & 2013-01 & 700 & 65 & 41 & 1.21 & 1 & C & C \\
2 & 2013-02 & 700 & 64 & 42 & 1.22 & 1 & C & C \\
2 & 2013-03 & 700 & 63 & 43 & 1.25 & 1 & C & C \\
2 & 2013-04 & 700 & 62 & 44 & 1.18 & 1 & C & C \\
2 & 2013-05 & 700 & 61 & 45 & 1.22 & 1 & C & C \\
\hline
\end{tabular}
\end{center}
\end{table}

\paragraph{}

The fields presented here are a handful of the more important factors driving mortgage behavior. A complete mortgage model would include dozens of effects, but it is not helpful to list them all out here. 

\begin{itemize}
\item loanId: A unique ID identifying the loan.
\item month: The month of the observation. 
\item FICO: The borrower's FICO score at origination. 
\item MTMLTV: The LTV of the loan modified for house price, amortization and curtailment. 
\item age: The age of the loan, in months.
\item incentive: The interest rate improvement the borrower would expect from refinancing. 
\item isOwner: Is this home occupied by its owner.
\item startStatus: What was the status of the loan at month start.
\item endStatus: What was the status of the loan at month end.
\end{itemize}

\paragraph{}

Each row in the table above represents one observation. A given loan would be represented by many observations. Some of the columns (such as FICO) that don't change with time will depend only on the loanId, others will depend on loanId and month. Computing several of these columns could be a very involved process. For instance, MTMLTV requires knowing the value of the home as well as any amortization or curtailment undertaken by the borrower. Incentive requires an estimate of what rate the borrower would be offered if he decided to refi in the given month. When fitting on historical data, these calculations are often much easier, as the historical data is known so fewer sub-models are needed to make forecasts. 

\paragraph{}

For the purposes of this paper, it is not important to understand the specifics behind these regressors, other than to know a few salient facts. 

\begin{enumerate}
\item There are a large number of them.
\item Some of these regressors require substantial calculation to derive. 
\item Some of these regressors are highly path dependent, depending on previous status transitions. 
\end{enumerate}

\paragraph{}

\section{Efficiency Considerations}

\paragraph{}

For the space of models in question, efficiency will be very important. When discussed here, there are two types of efficiency considered. 

\begin{itemize}
\item Computational Efficiency
\item Information Theoretic Efficiency
\end{itemize}

\paragraph{}

A good model in this space is one that very closely approximates the actual phenomenon under consideration using minimal human intervention and a reasonable amount of computing resources.

\subsection{Computational Efficiency}

\paragraph{}

Within the world of mortgage modeling, the computational space is huge. There are roughly 60 million mortgages in the country. Mortgage models typically require path dependent effects. As just one example, one of the best predictors for whether or not a borrower will miss a payment is the number of months since the last missed payment. Regressors such as this ensure that while each loan-month transition probability may have a closed form, there is no (known) closed form for the distribution of loan status at any time more than 1 month in the future. Therefore, if loan level accuracy is desired, then numerous (e.g. 1000+) simulations will be needed. A typical loan (on a typical path) will require an average of approximately 100 months of projection before liquidation, since the typical mortgage is refinanced approximately once every 5-10 years. In addition, the financial firms that uses these models typically need to examine a significant number (e.g. 20) of interest rate and house price scenarios. So the number of loan-months we would like to compute each day is: 

\begin{itemize}
\item $ n_{loan} = 6*10^7 $
\item $ n_{path} = 10^{4} $
\item $ n_{month} = 10^2 $
\item $ n_{scenario} = 2*10^{1} $
\end{itemize}

\paragraph{}

\begin{equation}
n_{loanMonth} = n_{loan} * n_{path} * n_{month} * n_{scenario} = 1.2*10^{14}
\end{equation}

\paragraph{}

Given that a single call to $ e^{x} $ takes on the order of 100 CPU clock cycles, assume that a model takes at least 10,000 cycles per loan-month. This implies that computing the whole universe would take at least $ 1.2*10^{18} $ clock cycles. This is about 500 million CPU seconds, or approximately 110,000 CPU hours. Clearly, running on a compute grid will be necessary to perform any meaningful fraction of this computation. 

\paragraph{}

Throughout this paper, it will be important to consider the computational cost of the functions being used. Generally, only entire (in the sense of complex analysis) functions should be used, as they are highly suitable for numerical optimization. Being entire, they are defined everywhere (so there is no "edge" for the optimizer to hit), and they have infinitely many continuous derivatives, ensuring that they are smooth enough to be optimized easily. Of these functions, polynomials are typically cheaper to compute than other functions in this family. Unfortunately, polynomials are often unsuitable due to issues with oscillations, closure, and unboundedness. Of the non-polynomial entire functions, the exponential function is unusually inexpensive, and therefore will be used in preference to other options whenever it is appropriate. 

\paragraph{}

It is these efficiency considerations, as much as the related explainability considerations that lead the models chosen in this field to be overwhelmingly parametric.

\subsection{Information Theoretic Efficiency}

\paragraph{}

Within this space, the data itself is not believed to be generated by any exact closed form formula. Therefore, it is important that the family of functions making up the model be able to approximate the physical phenomenon in question very closely. Appendix II has some definitions and discussion of these factors. 

\paragraph{}

The important point to take away from Appendix II is that most models will not converge to the exact distribution of the data. In this problem space, the dataset is large, so if the model fitting itself is efficient in the sense that the variance of the parameters is small, then the error in the model will come to be dominated by the mismatch between the model form itself, and the distribution of the data. The primary purpose of ITEM is to eliminate some of this mismatch, and thus produce a more accurate model. 

\paragraph{}

If a distribution is the limit of a sequence of models, it is clear that in realistic cases it is the limit of such a sequence as the number of parameters goes to infinity. If it were a limit of a sequence of functions $ g(\vec x | \theta) $ with a fixed set of parameters $ \theta $, then it would be equal to $ g(\vec x | \theta_{MLE}) $, which is by assumption not the case. So it is important that this sequence converge quickly, without including an unnecessarily large number of parameters. In future work, this question could be explored more completely in the context of Kolmogorov Complexity, but for the purposes of this paper it is enough to have heuristic means by which to prevent the unnecessary waste of parameters.

\section{Model Choice}

\paragraph{}

There are several broad families of models that could be applied to this space. A sample is explored below. 

\begin{enumerate}
\item Linear Regression
\item Nonlinear Regression
\item Kernel Smoothing
\item Neural Network
\item Random Forest
\end{enumerate}

\paragraph{}

In this problem space, the predicted quantity in question is a probability. Therefore, linear regression is totally unsuitable, as it can give rise to probabilities that are greater than 1.0 or less than 0.0. Even if this were to be bounded, it doesn't have the expected saturation behavior. We would expect that whatever function we choose would steadily approach some level of probability as the regressors become more or less favorable. 

\paragraph{}

Computational cost and engineering considerations eliminate kernel smoothing and neural networks from consideration. Neural networks themselves would blow far beyond the 10,000 CPU cycle computational budget envisioned for this model. In addition, there are issues with explainability. Kernel smoothing suffers from the correlation among loans. Due to macroeconomic variables, the entire loan population is moving with time, so most loans in the future will not be very near to any significant number of loans from the past. There are also issues with tiling this dataset properly to be able to run it on a grid, and with the fairly high dimensionality of the data itself. The space is simply too large (and too mobile) for kernel smoothing to work well. 

\paragraph{}

Lastly, the random forest could work, but it has issues with explainability and also parameter efficiency. Simply put, the random forest will use far too many parameters, and run the risk of very dangerous overfitting. With regards to explainability, in the mortgage world, people talk about the "credit box", which is in typical parameter space, quite literally a box. For instance, (LTV $< $ 80, Fico $> $ 720) is a reasonable credit box. Even defining structures such as this becomes very hard with a random forest approach, requiring extensive simulation. For something like multinomial logistic, it is pretty easy (though not trivial) to define level sets of the function. These level sets may be high dimensional, so some projection is requied in order to get anything that is very explainable. In the absence of the need for very high precision, a credit box approach can serve as a simple substitue. 

\paragraph{}

For all of these reasons, modern mortgage models are typically parametric.

\subsection{Parametric Models}

Within mortgage modeling, there are two widely used  families of parametric models. 

\begin{enumerate}
\item Structural models
\item Multinomial regressions
\end{enumerate}

\paragraph{}

There are advantages and disadvantages of each which will be discussed in turn. 

\subsubsection{Structural Models}

\paragraph{}

For a good overview of a structural model, see \cite{JPM}. When modelers talk of structural models, they typically mean a model that is not the result of a maximum likelihood optimization. Occasionally, (as in \cite{JPM}) a structural model is more explicitly defined as hedonic (see \cite{HED}) model driven by known (e.g. economic) factors. Usually a true structural model is both of these things, and will take the form of a collection of unrelated terms that are estimated through manual examination of the dataset, or by drawing some analogy between the data and relevant policy or law. The problem with this approach is that it is hugely expensive in terms of man hours, tends to have explainability issues (i.e. why did you add this rather than that), and almost always results in parameters that are not very optimal. It doesn't help that only a handful of effects can be discovered through this manual process, and people are not good at visualizing data in higher dimensions. 

\paragraph{}

Structural models are effective when the data is very limited, and when there is insufficient computational power to fit an accurate model on a large dataset. Historically, this was very much the case, which is why older mortgage models tend to be structural. However, when the dataset size is very large and computational resources are readily available for fitting, structural models tend to be less efficient than nonlinear regressiosn both in terms of information theory and in terms of computational performance. 

\paragraph{}

This inefficiency is caused by several factors. The first is that (by our definition here), in a structural model, the parameters are provided by hand and are not the result of an MLE estimation. It is certainly possible that $ \theta_{structural} = \theta_{MLE} $, but in practice it is unlikely. For this reason, the model will almost always be biased, and the variance of the parameter estimates will be needlessly high. In addition, it is possible that the parameters are added in an efficiently parsimonious fashion so as to not overparameterize the model, but again, without carefully searching the problem space, this is unlikely to happen of its own accord in practice. In addition, it is certainly possible that the model could be constructed from entire functions chosen for efficiency, but again, this is not common in practice. As a result, these models tend to be inaccurate, bloated with excess parameters, and computationall inefficient.

\subsubsection{Multinomial Regression}

\paragraph{}

A nonlinear regression needs to properly represent the problem space. Primarily, this means that it needs to produce a probability vector for any input. A broad class of functions that fit this description is the class of multinomial cumulative distribution functions. This family of functions takes a vector of values (often referred to as "power scores" or "propensity scores") as input, and produces a probability vector as output. Common examples of this family are logistic (associated with the logit function), and the gaussian CDF (associated with the probit function). 

\paragraph{}

For these models, if there is a strong belief that the data actually follows one of these distributions, then this should certainly be considered. For most datasets, and particularly for mortgages, there is no reason to believe that one of these distributions is a more natural fit than the other. For this reason, the logit model is typically used because it is computationally cheaper than the probit model.

\subsection{Multinomial Logistic Regression}

Since modern mortgage modeling is typically based on multinomial logistic regression, it is helpful to review the technique here. One can see an overview of the multinomial logistic regression model in \cite{CZEPIEL}.

\paragraph{}

Consider first the function multinomial logistic function. 

\begin{equation}
L(\vec v) = \frac{ e^{\vec v} }{| e^{\vec v}|}
\end{equation}

\paragraph{}

It is understood in the definition above that the exponential of a vector is taken element wise to produce a vector, and that $ \beta $ is a matrix of parameters. The values of $ \vec v $ are known as power scores. If we set the probability vector $ \vec y = L(\vec v) $, then this equation is underdetermined since the values of $ \vec y $ sum to 1. The equivalent invariant for $ \vec v $ is that subtracting the same value from all elements of $ \vec v $ leaves the values of $ \vec y $ unchanged. Traditionally, the largest power score is subtracted from all the power scores, guaranteeing one power score is $ 0 $ and the rest are negative. This improves the numerical stability of the logistic function itself and ensures that overflow results in the correct limiting behavior rather than NaN. With that convention in place, $ L(\vec v) $ defines a bijection, so it has an inverse. 

\begin{equation}
L^{-1}(\vec y) = \ln(\frac{\vec y}{y_{0}})
\end{equation}

Where here $ y_{0} $ is the probability corresponding to the $ 0 $ element of $ \vec v $. These power scores are then derrived from the regressors using a simple dot product. 

\begin{equation}
\vec v = \beta \cdot \vec x
\end{equation}

Modeling the data in this way makes the following assumptions. 

\begin{itemize}
\item $ \vec y $ is linear with respect to $ \vec x $ in logit space. 
\item The terms of $ \vec x $ interact through summation in logit space. 
\end{itemize}

\paragraph{}

These limitations are critically important, in that they are typically not satisfied by any real world phenomenon. This problem is resolved by introducing an arbitrary vector valued function $ F(\vec x) $ that transforms the values of $ \vec x $ into new regressors which are linear with respect to $ \vec y $ in logit space and do interact by summation.

\paragraph{}

Now the model looks like this. 

\begin{equation}
L(\beta \cdot F(\vec x_i)) = \frac{ e^{\beta \cdot F(\vec x_i)} }{| e^{\beta \cdot F(\vec x_i)}|}
\end{equation}

This equation is now perfectly general. Any probability vector can be written in such a form provided $ F(\vec x) $ can be discovered, and also that one is willing to allow $ \beta $ to have entries in the generalized reals $ ( \mathbb{R} \cup \{+\infty, -\infty \} )  $.

\paragraph{}

This function has a (relatively) simple closed form derivative. Start by computing the derivative with respect to the power score. 

\paragraph{}

\begin{equation}
\frac{d}{ds_{i}} L(\vec s)_{w} = (\delta_{i, w} - L(\vec s)_{i}) L(\vec s)_{w} ds_{i}
\end{equation}

\paragraph{}

Now if $ s_{i} = \beta_{i} \cdot F(\vec x) $ then the derivative with respect to $ \beta_{i, j} $ follows. 

\paragraph{}

\begin{equation}
\frac{d}{d \beta_{i,k}} L(\beta \cdot F(\vec x))_{w} = (\delta_{i, w} - L(\beta \cdot F(\vec x))_{i}) L(\beta \cdot F(\vec x))_{w} F(\vec x)_{k}
\end{equation}

\paragraph{}

Similarly, if $ F(\vec x) $ is parameterized by $ \vec \alpha $, then 

\paragraph{}

\begin{equation}
\frac{d}{d \alpha_{i}} L(\beta \cdot F_{\vec \alpha}(\vec x))_{w} = (\delta_{i, w} - L(\beta \cdot F_{\vec \alpha}(\vec x))_{i}) L(\beta \cdot F(\vec x))_{w}(\beta_{w} \cdot \frac{d}{d \alpha_{i}}F_{\vec \alpha}(\vec x)_{k})
\end{equation}

\paragraph{}

Likewise, the derivative of the log likelihood can be computed. 

\paragraph{}

\begin{equation}
\frac{d}{ds_{i}} -\vec y \cdot \ln(L(\vec s)) = \sum_{w} y_{w} * (\delta_{i, w} - L(\vec s)_{i}) ds_{i}
\end{equation}

In the mortgage modeling world (returning to the example), the vector $ \vec x_{i} $ is the data for the i'th loan-month, and the vector $ \vec y_{i} $ represents the probability that the loan will be in each of the available loan statuses in the next month. The parameters $ \beta $ depend on the starting state, for this example we can assume that the model under consideration is for computing the $ C \rightarrow * $ transitions. The loan-months are then simply partitioned by starting state, and the appropriate value of $ \beta $ used for each. 

\paragraph{}

The difficulty now is reduced to finding a suitable value of $ F $. The model would be able to converge to any distribution of $ \vec y $ if the corresponding value of $ F $ could be found. If no such value of $ F $ can be found, then the model will converge to a distribution that is a non-infinitesimal distance (in KL divergence, for instance) from the true distribution of $ \vec y $. Given the function $ F $, the remaining fitting of the model is an entirely mechanical maximum likelihood estimation.

\subsection{Optimality of Logistic Regression}

\paragraph{}

The logistic regression model is extremely efficient for several reasons. First of all, it is critical that only analytic functions be considered. Analytic functions are always $ C^{\infty} $, and thus it is always possible for an automatic nonlinear optimizer to solve for its parameters. In addition, analytic functions are closed under composition, so it is easy to build analytic functions using simpler analytic functions as building blocks. It should be possible to make a good model using $ C^{\infty} $ functions that are not analytic, but such functions are rare and often difficult to construct. As a practical matter, all reasonable functions that are $ C^{\infty} $ are also analytic, so this is a distinction without a meaningful difference. 

\paragraph{}

Multinomial logistic regression is analytic, and it also has computational efficiency that is in some sense optimal. See Appendix I for more details. In brief, any analytic CDF will be based on one or more analytic functions that are not polynomials. The computationally cheapest such function (subject to certain conditions) is the exponential function. Therefore a simple formula incorporating exponentials will be more computationally efficient than any other option. The logistic function is just such a function. In addition to the above mentioned traits, the logistic function has arbitrarily many simple closed form derivatives. This greatly accelerates the process of fitting model parameters and reduces the associated errors. 

\paragraph{}

Lastly, if the proper value for $ F $ can be discovered, then the logistic regression will converge arbitrarily closely to any distribution. In that sense, it is completely general. In addition, if the function $ F $ can be recovered efficiently from the data at hand, then this regression will also be information theoretically efficient in that it will produce parameters with low variance and converge quickly towards the actual data distribution, thus producing a model with small residuals. The ITEM system attempts to recover this value of $ F $ automatically.

\subsection{Splined Multinomial Logistic Regression}

\paragraph{}

Returning to the mortgage modeling problem described in Section \ref{example}.

\paragraph{}

For the multinomial logistic models, the problem usually begins with variable selection. Here, expert knowledge is important. Some of the regressors are fairly obvious, but others reflect real economic motives and drives, but are only obvious when examining certain interactions between the dataset and historical data from other sources. In any case, a candidate list of regressors is drawn up. It is helpful to eliminate regressors that do not have much impact, and often modelers do so using p-tests and Lasso style regularization to come to an initial fit.  Initially, the function $ F $ is simply taken to be the identity function, or perhaps some simple splines chosen a-priori.

\paragraph{}

Once this is done, the next step is to look at the residuals. Observing a typical set of residuals, it will generally be the case that there is some smooth deviation from the projection. It appears that the projection is following one curve, but the actual data is following another. The mean of the two (for a suitable definition of mean) must be the same, but the curves do not in general coincide. This situation comes about when the response to a variable is not linear in logit space. 

\paragraph{}

The solution to this problem is generally to construct some splines, the linear step spline defined below is typical.

\begin{equation}
s_{a, b}(x) = min(1, max(0, \frac{x-a}{b-a}))
\end{equation}

\paragraph{}

Then the model would be recalculated using this spline instead of (or perhaps in addition to) the original regressor. 

\begin{equation}
L( ... ) = \frac{e^{\beta_{0} + \beta_{1} \cdot s_{a, b}(x_{1}) + ...}}{| ... |}
\end{equation}

\paragraph{}

By adding several of these splines, the response curve for $ x_{1} $ could slowly be transformed into any particular curve, making the approximation better and better. In addition, this spline is bounded, so it naturally defends the model against extreme outliers. 

\paragraph{}

This spline is an example of a (non-analytic) CDF. Unfortunately, because this curve is not analytic with respect to $ a $ and $ b $, it is usually not possible to run an optimizer over it in order to find the optimal values of these parameters. Therefore, the values of $ a $ and $ b $ are typically chosen manually through inspection of the data and model residuals. This is extremely expensive in terms of man-hours, and results in splines that are almost never truly optimal. In addition, when adding splines, it is not typically feasible to determine which regressor most needs a spline, so often the splines actually included (even with optimal parameters) are less helpful than other splines which were not included. Using a sequence of these splines, the model can approach the true distribution arbitrarily closely, but it cannot do so without almost unlimited manual labor.

\paragraph{}

Often, modeling groups attempt to avoid this unlimited investment of manual effort by automatically fitting splines. Typically, before the process even begins it has failed. Step splines such as this cannot be fit automatically (with typical algorithms at least) because they are not analytical in their parameters. Often, this gives rise to an attemp to use cubic splines (which suffer from the same problem) or polynomials. 

\paragraph{}

Polynomials deserve special mention, because they are analytic. However, just the same, they will not work. First of all, polynomials are never bounded, so even a few outliers will completely destroy the fitting process. Secondly, polynomials of degree $ d $ make a closed group, adding more polynomials of the same or lesser degree simply produces another polynomial of the same degree. Therefore, in order to add "more" curves, it is necessary to instead add polynomials of progressively higher and higher degree, which brings us to the next issue. Polynomials are strongly oscillitory, having all manner of harmonics and "ringing" issues which rapidly make the model nonsensical. Polynomial ringing is the wild oscillitory behavior caused by small moves in two points that are very close together and that the polynomial is constrained to pass through. 

\paragraph{}

When this work is tied to the notion that these response curves need to be splines or polynomials, there is never any real hope of success. 

\paragraph{}

With an algorithm that can select proper curves (not necessarily splines) automatically, the model would be much improved. This is the goal of ITEM.

\section{An Introduction to ITEM}

The ITEM model builds on the splined multinomial logistic regression by providing a means to fit splines automatically. To get good results, the family of curves chosen must meet the following minimum critera. 

\begin{enumerate}
\item The family must be defined by a few (e.g. 2) parameters
\item It must be bounded and entire in all inputs (parameters as well as $ x $).
\item It must be very smooth and have at most one local maximum internal to the dataset (i.e. no harmonics)
\item The family must form a complete basis. Ideally, realistic curves are well approximated by a few members
\item It must be efficient to calculate
\item It must have an efficient closed form derivative.
\end{enumerate}

\paragraph{}

The first requirement is necessary for information-theoretic reasons. If the model is to use the AIC to determine which curves to include, this will be greatly undermined if the curves require a huge number of parameters. In addition, the optimizers that will make these curves will have a much harder time finding good results.

\paragraph{}

The second requirement (together with the first) is basically the same as saying that an optimizer can be used to find the parameters. Recall that a curve is entire if it is analytic and defined for all points. For this purpose, we will consider a curve to be entire if it is defined on the whole real line. Similarly, no properly entire function can be bounded (Liouville's Theorem), but one could be bounded on the real line, which is the important factor here. 

\paragraph{}

The analytic requirement ensures that the curve has well defined derivatives. The boundedness and requirement that the function be entire protect against computational problems when an unrealistic starting point is given to the optimizer. The requirement that it be entire also ensures that there are no large zones where the derivative is zero (such as in the spline case). For instance, when optimizing a spline or similar curve, if both $ a $ and $ b $ are chosen to be below the whole dataset, the optimizer will find all progress impossible because the derivative is identically zero. Choosing curves bounded at 1 in particular helps to make the associated values of $ \beta $ easier to interpret. 

\paragraph{}

In addition to the computational utility of using analytic functions, there is a physical justification as well. In most realistic datasets, there will be some degree of measurement error. If these measurement errors are (for instance) gaussian, then any properly specified model must be composed of only analytic response functions. We know this because the model response to the data must be a convolution between the error function and the actual functional form of the underlying phenomenon. Analytic functions are contagious over convolutions, convolving an analytic function with any other reasonable (e.g. bounded with finitely many discontinuities) function produces an analytic function. Therefore, any model operating on a dataset with analytic (again, typically gaussian) measurement errors is automatically misspecified if it includes any unanalytic response functions. This is one reason to avoid splines, the sharp turns they make are simply not possible if the visible data has any noise relative to the state driving the modeled phenomenon. 

\paragraph{}

The third requirement ensures that results for similar inputs will be similar, a basic requirement for any model. 

\paragraph{}

The fourth requirement ensures that any curve can be suitably approximated by a sum of enough curves from this family. This is necessary if the model is to actually converge to the true distribution for large dataset sizes. 

\paragraph{}

The fifth requirement is a basic requirement that the model shouldn't be wasting money, once all the other requirements are satisfied. 

\paragraph{}

Most of the curves that modelers try to use here fail at least one of these tests. Requirement 1 eliminates kernel smoothing, requirement 2 eliminates splines, requirmements 3 and 4 together with 1 eliminate polynomials.

\subsection{ITEM Curve Families}

\paragraph{}

Notice that for any cumulative distribution function $ CDF(x) $ that is analytic in $ x $, the function $ CDF(a^{2}(x-b)) $ will satisfy all requirements, provided it is efficient to calculate. The logistic function is an efficent CDF, and it has closed form derivatives, so that is a natural choice. The value of $ a^{2} $ is used so that this CDF is always upwards sloping, hence the associated $ \beta $ has the expected sign, positive for positive correlation, negative for negative correlation. This function also has closed form derivatives, which make it much easier to handle within the optimizer. 

\paragraph{}

The formula is then 

\begin{equation}
C_{a, b}(x) = \frac{1}{1 + e^{-a^{2}(x-b)}}
\end{equation}

The derivatives are

\begin{equation}
\frac{d}{da} C_{a, b}(x) = 2a(x-b)C_{a, b}(x)C_{a, b}(-x)
\end{equation}

and

\begin{equation}
\frac{d}{db} C_{a, b}(x) = -a^{2}C_{a, b}(x)C_{a, b}(-x)
\end{equation}

\paragraph{}

Unfortunately, this function family is not enough to achieve good results. The problem comes down to the procedure for doing the fit itself. In the case of a regressor with a response that is strongly peaked within the distribution (e.g. it looks like a gaussian), no single CDF can improve the results significantly. Two of them taken together could approximate a gaussian, but if the fitting procedure is adding curves one at a time, this does not help. Therefore, it is necessary to include a spike-like family of curves as well. Fairly obviously, the spike family should be actual gaussians. 

\begin{equation}
G_{a, b}(x) = e^{\frac{-(x-a)^{2}}{2b^{2}}}
\end{equation}

Its derivatives are

\begin{equation}
G_{a, b}(x) = \frac{2(x-a)}{2b^{2}} G_{a,b}(x)
\end{equation}

and

\begin{equation}
G_{a, b}(x) = \frac{2(x-a)^{2}}{2b^{3}} G_{a,b}(x)
\end{equation}

\paragraph{}

Given that the gaussian family is used, if the centrality parameter of the gaussian is set outside of the bulk of the dataset, the result would be a bounded monotonically increasing function within the dataset. It then stands to reason that this might eliminate the need for the Logistic family. Unfortunately, this is not the case. The Gaussian family will tend to have a very large derivative near the edge of the dataset, whereas many features of datasets like the one in the example problem saturate rapidly well within the range covered by the data, and have a much more classic S-curve shape. Therefore the Gaussian family itself will yield poor results, and it is worth keeping the Logistic family as well. 

With these families in hand, it is then necessary to fit them. 

\subsection{Comparison to Generalized Additive Models}

\paragraph{}

ITEM may be considered to be a special case of a Generalized Additive Model \cite{GAM}, albeit one that is parametric. Since the nature of the problem has eliminated from consideration all nonparametric models, a proper nonparametric GAM could not be considered for this problem space. The issue is that in order to make any predictions, a true GAM would require the whole dataset resident in memory, this is not realistic for multi-TB data sets.  

\paragraph{}

A slight modification of a GAM could be used, in that the data for each parameter could be bucketed and averaged, then these averages connected with (for instance) a cubic spline. This would have the advantage of being comparatively quick to fit and also relatively efficient to use. The disadvantage is that if the bucket size is too small, there will be a lot of oscillation, but a bucket size too large will lose a lot of detail. An algorithm similar to that used by ITEM to determine the number of buckets using an information criterion could be used, but that would considerably slow the fitting. There are also many potential issues related to the fact that for typical problems, highly non-uniform buckets are likely to be called for, greatly expanding the needed number of parameters. Some of these parameters would then be knot points, which would potentially have numerous issues due to unanalytic results when the knots get close together. It remains to be seen whether a bucketing approach such as this could achieve information theoretic efficiency on par with the ITEM curves themselves. Where there is a strong suspicion that the data follows an S-curve, for instance, ITEM would be expected to have a significant advantage. It can closely approximate an S-curve with just two parameters, whereas a more typical bucketed GAM would require many.

\subsection{Information Theoretical Considerations}

\paragraph{}

Once an automated procedure is unleashed upon curve selection, it is critically important to handle two information-theoretical considerations. 

\subsubsection{Efficient Convergence}

\paragraph{}

The model must efficiently converge to the true distribution of the dataset. The functions chosen above were selected such that most typical data distributions will be well approximated by a small number of functions that can be selected one at a time. The choice of functions changes the assumptions related to our model from the classic logistic model pair to a new set of greatly relaxed assumptions. 

\begin{itemize}
\item The function $ L^{-1}(\vec y) $ is not pathologicial (* see below). 
\item The terms of $ \vec x $ interact through summation in logit space. 
\end{itemize}

\paragraph{}

Notice the first condition. By "not pathological" we really mean just that it is efficiently approximated by a limited number of logistic and gaussian functions. For instance the condition that its Fourier transform decays rapidly in frequency is sufficient. Rather than requiring linearity, we require only that the function is not pathological, a much weaker condition. The second condition remains unchanged. If there are complicated interaction terms between the variables, it will still be up to a human operator to discover that. No similar scheme can be attempted to uncover interaction terms since the number of such terms at each step is equal to the power set of the elements of x times the number of interaction functions, a space much too large to search effectively and posessing no obvious metrics to allow an optimization. 

\paragraph{}

However, subject to these assumptions, the model will now converge to the true distribution if given enough parameters.

\subsubsection{Resistance to Overfitting}

\paragraph{}

To achieve optimal accuracy, the model must not incorporate curves that are not well supported by the dataset. This inclusion of extraneous features is called overfitting. This requirement has two portions. 

\begin{enumerate}
\item The model should stop calculating when there is no point in further work
\item The model must not include features that are not well supported by the dataset
\end{enumerate}

\paragraph{}

The key here is to use an information criterion. At each stage of the fitting, include the best available curve, but demand that it pass an information criterion before allowing it to be included into the model. When the best curve can no longer pass the criterion, the model is complete. In the case of ITEM, this is done using the AIC/BIC. These criteria differ by only a constant of $ (2 - \ln(N))k $, with the BIC being the stricter criterion. In the results section, the choice of stopping condition is discussed. Typically, it won't matter, an AIC criterion may include marginally more curves than BIC, but the difference will be small. In many tests, they include the exact same curves. There is some argument to be made that the BIC could be used simply because it is somewhat computationally cheaper since it may stop drawing curves earlier, and it is better to err on the side of fewer rather than more curves when in doubt.

\section{Implementation of ITEM}

There are many subtle points in the actual fitting of the ITEM model, so that procedure will be described here.

\subsection{Fitting ITEM curves}

Regressors can be divided into two types. 

\begin{itemize}
\item Binary Flags
\item Real Numbers
\end{itemize}

\paragraph{}

Binary flags are just what they sound like, being either $ 1 $ or $ 0 $. They don't need to be further considered here, since they have no need of curves, a simple fit to produce a corresponding $ \beta $ will suffice. 

\paragraph{}

Real numbers are typically double precision IEEE floating point numbers. It is important that these values have a well defined (i.e. not discrete) metric. A discrete metric simply means that the regressor is categorical, see \cite{DISCRETE}. Having a non-discrete metric over a space means that there is some sensible notion of distance. Given such a notion, it should be possible to approximate the apparent power score of the data (as evidenced by the observed transitions) by a smooth curve. The exact shape of the dataset itself is not important, but it is important that it (for example) have a fourier transform with rapidly decaying high frequencies. For instance, it should not be oscillatory with a high frequency. These pathologies are not common. 

\paragraph{}

If a regressor happens to be a categorical variable (i.e. it is drawn from an enumeration of $ N $ values, but without a well defined metric), then it should be replaced with $ N - 1 $ flags. It is important to not form a complete basis with any flags that are introduced, as that would cause a linear correlation between the sum of the flags (guaranteed to be 1) and the intercept term of the regression (also guaranteed to be 1). 

\paragraph{}

For numerical stability, it is very important to cap the log likelihood at some maximum constant. Approximately 20 is reasonable. This prevents the case where a specific loan with extremely odd parameters drives the entire fit by producing a vanishingly low probability for an even that actually occurs in the dataset. Adding a small random noise (see Annealing, \ref{annealing}) can also help here. 

\paragraph{}

The basic fitting procedure is as follows. 

\begin{enumerate}
\item Fit all flags and betas for existing curves.
\item If model has maximum number of curves, exit. 
\item For each regressor
\item For each transition
\item For each curve type (i.e. logistic, gaussian)
\item Use an optimizer to find the curve that minimizes the log likelihood.
\item Consider all curves thus computed, take the on that reduces the log likelihood the most. 
\item For this best curve, examine the AIC, verify that the test passes.
\item If AIC test fails, exit, otherwise add the curve and go to step 1
\end{enumerate}

\paragraph{}

Notice that it is critically important that the curve families were chosen such that a single curve will always improve the fit. This ensures that the optimization procedure above can always make progress and will not get stuck in a situation where fitting two curves at once would succeed, except in the most marginal cases. 

\paragraph{}

During the fitting, there are a few pitfalls. The first thing to remember is that the logistic function has an inverse. Therefore, it is most efficient to convert the model probabilities into power scores, then do an (N + 2) parameter fit where N is the number of parameters in the curve (i.e, 2 for the curves mentioned above). This will be fitting the 2 curve parameters, the curve beta, and an intercept adjustment. All 4 are needed in order to arrive at the actual optimum. For instance, without fitting the intercept adjustment, it will typically be impossible to draw any curve that improves the fit, since any curve will increase or decrease the population-wide average, making the fit worse. 

\paragraph{}

Once the power scores are availble, on each iteration simply update $ score_{i} = score_{i} + intercept + \beta*C_{a, b}(x) $. This will require only the regressor being fit against and the 4 parameters, it will not require the rest of the regressors which would otherwise bog down the calculation. Then convert these scores back to probabilities, and compute the log likelihood. Remember to use the analytical derivatives. Most of these calculations will not need to process the whole dataset, see the section on adaptive optimzation (Adaptive Optimization \ref{adaptive}). 

\paragraph{}

Each curve so fit will cost 3 parameters, 2 for the curve, 1 for the beta, the intercept is an adjustment of an existing parameter, so it is free. These curves can be made available to other transitions at a cost of 1 parameter each, and they could even be chosen such that when applied across all transitions the model is most improved. Both of these paths should be avoided, as either one will result in a model that is less understandable, though admittedly the information-theoretic efficiency may be marginally improved. When sharing curves like this, a few parameters are avoided, but each transition now gets a curve that is not ideal for it, and it may be hard to explain why some features from one transition are suddenly showing up in another. This is a judgement call, and the one place where efficiency has been sacrificed for interpretability within the model.

\subsection{Annealing ITEM curves}
\label{annealing}

\paragraph{}

Once all curves are fit, and there is either insufficient information to fit further curves or the model has reached the operator supplied curve limit, the model is complete. At this point, however, the model may not actually be optimal. The exact curves chosen are path dependent, and may not be a global optimum. 

\paragraph{}

This is analogous to metal that has been formed and then cooled. If it was cooled quickly, the internal arrangement of atoms may not be at the minimum energy. Essentially, the crystal was left at a local minimum somewhere along the way to the global minimum. Given more time (and sufficient energy), it could eventually, through random chance, bounce out of the local minimum and continue towards the global minimum. At low temperatures, there is not enough energy for this to occur in any reasonable timeframe. The solution in metalurgy is called annealing, simply heat the metal and then cool it very slowly. There is a procedure (\cite{ANNEALING}) that can improve the total fit quality at the cost of computational time during the fit.

\paragraph{}

This pass is usually not necessary, but if extra CPU cycles are available, it may be attempted. Dramatic improvements from this process should not be expected, but neither are dramatic changes. Most models go through many iterations of fitting and analysis (or bug fixing), so fitting performance is important. However, for a final version, it may be worthwhile to spend the extra resources to get a marginally better product. The fit below assumes that curve parameters are not optimized during step 1. From experience, adding so many parameters to the optimization rapidly bumps against machine precision issues and causes the results to be worse than an iterative approach.

\begin{enumerate}
\item Run a full optimization on all $ \beta $ values in the model (but not curve parameters). 
\item If loop count reached, exit. 
\item Loop over each regressor. 
\item Remove all the curves from this regressor, then (as above) generate the same number of new curves. 
\item Once all regressors have been processed in this way, go back to step (1). 
\end{enumerate}

\paragraph{}

For this to really work well, some small amount of randomness needs to be added to the model. In this way, any curves that came about simply because of the exact content of this dataset will have a chance to be replaced by other curves that may be better, but are unreachable from this dataset. The proposed method is to add to each observed probability a gaussian noise with a very small standard deviation (around 1.0e-4 to 1.0e-6 depending on dataset size), and then renormalize the results. This has the effect of making events that didn't happen (probability 0 in the dataset) have some small probability of occuring, and thus they cannot be completely discounted by the model. The size of this standard deviation should be inversely proportional to the dataset size. Essentially, in a dataset with a million points, nothing should be considered less than approximately a one in a million chance, for instance. The data itself simply does not support such a strong assertion. 

\paragraph{}

Another way of introducing randomness, is in the selection fo the starting points for the centrality parameters of the various curves to be drawn. This could be started at the actual dataset mean, but that may prevent improvement in some cases, where a local optimimum prevents the the global optimimum from being obtained. An example related to seasonality will be discussed in Section \ref{fitResult}. One way to avoid this is to instead select a random data point, and start the centrality parameter at that point's value. This has the effect of allowing some out of the way portions of the parameter space to be explored, but it does not tend to explore places lacking in data. Since the curve fitting process is iterative, there will be many chances to find a good curve, especially if annealing is used. For curves that have an easily reachable global optimum, this won't matter anyway. 

\paragraph{}

For the example built here, annealing results will not be discussed. In future work, good values of the noise term and annealing performance analysis can be discussed.

\subsection{Adaptive Optimization}
\label{adaptive}

For the ITEM model, all the optimizations are taken over the expected value of the log-likelihood of the dataset. These functions look like this. 

\begin{equation}
f(\beta) = \frac{1}{N} \sum_{k=0}^{N} \vec y_{k} \cdot \ln(\vec L(\beta \cdot F(\vec x_{k})))
\end{equation} 

In this equation, $ N $ is the number of observations, and $ L(\beta \cdot F(\vec x_{k})) $ is the model estimated value of the transition probabilities. Considering just the partial sums of this function.

\begin{equation}
f_{M}(\beta) = \frac{1}{M} \sum_{k=0}^{M} \vec y_{k} \cdot \ln(\beta \cdot F(\vec x_{k}))
\end{equation}

It is clear that for large $ M $, $ f_{M}(\beta) - f(\beta) $ is distributed as $ N(0, \frac{\sigma}{\sqrt{M}}) $ where $ \sigma $ is simply the standard deviation of the average element from the dataset. Optimizers are driven not by the value of $ f(\beta) $, but by the differences between these values, e.g. $ f(\beta_{1}) - f(\beta_{2}) $. The standard deviation of this difference in the case of partial sums may be computed directly, call it $ \sigma(\beta_{1}, \beta_{2}, M) $

First, define a term wise difference. 

\begin{equation}
d(\beta_{1}, \beta_{2}, k) = \vec y_{k} \cdot (\ln(\vec L(\beta_{1} \cdot \vec x_{k})) - \ln(\vec L(\beta \cdot \beta_{2}, \vec x_{k}))))
\end{equation}

Then the std. deviation of the partial sum can be defined as follows. 

\begin{equation}
\sigma(\beta_{1}, \beta_{2}, M) = \frac{1}{\sqrt{M}} \sqrt{(d(\beta_{1}, \beta_{2}, k) - \mu)^{2}}
\end{equation}

Where here, $ \mu $ can be taken to simply be the sample mean. Due to the sizes of the numbers typically encountered, this will present no problem. Then, for some constant $ C $, successive values of $ M $ (starting at $ M_{0} $) are considered until 

\begin{equation}
| f_{M}(\beta_{1}) - f_{M}(\beta_{2}) | > C \sigma(\beta_{1}, \beta_{2}, M)  
\end{equation}

\paragraph{}

Here the choice for the value of $ C $ reflects how certain the modeler would like to be that the new point is actually better than the old point. For values of $ C $ around 5 (the standard metric for proof in scientific circles) the odds of spurious point selection are very small, roughly $ 10^{-6} $. In any case, the model is not very sensitive to this parameter, setting it smaller doesn't help performance much, and setting it larger doesn't improve accuracy meaningfully. The value of $ M_{0} $ needs to be chosen large enough that the law of large numbers will take effect, and the data sample will be representative of the whole dataset. Again, the model is not very sensitive to this parameter. Making it approximately 1\% of the data set size is a reasonable setting if the dataset is large. Alternatively something of moderate size such as $ 10^{4} $ should work for most data sets. Again, setting this parameter very small doesn't improve the efficiency much, so if it is a moderately sized fraction of the whole dataset size, little is gained by reducing it further. 

\paragraph{}

Note that for each successive value of M, the previous results need not be recalculated, only the additional elements need to be considered. If the dataset is large, then it will almost never be necessary to examine the whole dataset. Only in the last few iterations of the optimizer will the comparisons be close enough together that the entire dataset must be considered. 

\paragraph{}

Furthermore, this gives a natural stopping condition. Whenever the points considered by the optimizer satisfy

\begin{equation}
| f_{N}(\beta_{1}) - f_{N}(\beta_{2}) | < \sigma(\beta_{1}, \beta_{2}, N)  
\end{equation}

\paragraph{}

Applied over the whole dataset $ N $, then there is simply no further progress possible, so the current results should be returned as-is. 

\paragraph{}

The end result of this algorithm is that for a constant value of x and y tolerance (i.e. stop optimizing when the parameter is bracked close enough to its optimum, etc...) the cost of optimization grows sub linearly. For very large dataset sizes, this cost would approach a constant, as the optimizer would never reach the end of the dataset for any iteration.

\subsection{Hybrid Projection}

\paragraph{}

Once a model is fit, it must be used to project future behavior. Though this process will not be covered in depth in this paper, an efficient means for performing this projection will be described. 

\subsubsection{Markov Matrix Multplication}

\paragraph{}

If the model contains $ m $ potential states, then at each time period, the $ m \times m $ Markov matrix at a time $ t $ (here denoted $ M(t) $) may be computed. Then the next state $ \vec s(t) $ may be computed from $ \vec s(t-1) $ through matrix multiplication. 

\begin{equation}
\label{matmult}
\vec s(t) = M(t-1) \vec s(t-1)
\end{equation}

\paragraph{}

This method has two primary disadvantages. 

\begin{enumerate}
\item It forbids the use of non-Markovian regressors
\item It wastes resources on computations for extremely rare states
\end{enumerate}

\paragraph{}

In mortgage modeling, both of these disadvantages are  particularly severe. In general, a regressor such as "number of months since last delinquent" is incredibly important, but non-Markovian. In addition, some states (e.g. "In Foreclosure") are extremely rare. If the matrix multiplication method is used, then the majority of the computations each month go towards states that are immaterial for most loans, on most interest rate paths. This method is especially expensive since interest rates and housing prices need to be simulated anyway, so it doesn't even eliminate the need for simulation. 

\subsubsection{Simulation}

\paragraph{}

An alternative to matrix multiplication is to each month select a random transition to make, weighted by transition probability. Now instead of $ \vec s(t) $, the equation includes $ \vec {s'}(t) $ where $ \vec {s'}(t) $ is composed of a single $ 1 $ and all other elements are zero. This means that only one column of $ M(t) $ needs to be calculated. Call this selected matrix $ M'(t) $, it is composed of all zeros except for a single $ 1 $ element in the randomly selected (weighted by probability) location along the column corresponding to the $ 1 $ element in $ \vec{s'}(t-1) $. Then $ \vec {s'}(t) $ is now especially easy to calculate. 

\begin{equation}
\vec {s'}(t) = M'(t-1) \vec {s'}(t-1)
\end{equation}

\paragraph{}

This computation involves only a single column of $ M(t-1) $, and is thus much cheaper (generally about $ m $ times cheaper) than the computation in equation \ref{matmult}. Furthermore, since there is no longer any uncertainty about the loan status in any month, non-Markovian regressors no longer pose a problem. Unfortunately, this method introduces substantial additional simulation noise, which may require the use of additional paths, thus increasing computational cost. 

\paragraph{}

In typical usage, especially if loan level noise is not a problem (i.e. only aggregate behavior of a pool is needed), simulation is several times cheaper than matrix multiplication, even when it requires the use of extra paths due to excessive noise. This is especially true since many of the most expensive computations are related to some of the rarest states (e.g. "Foreclosure") in many loan level models.

\subsubsection{Hybrid Simulation}

\paragraph{}

Ideally, one would like a mechanism as inexpensive as simulation, but with as little noise as matrix multiplication. In addition, this method should allow the use of non-Markovian regressors, just like simulation does. 

\paragraph{}

Returning again to the mortgage modeling example, notice that most loans will be in status $ C $ most of the time. For any loan that starts in status $ C $, we could assume that it will always remain in $ C $. If we do so, then the non-Markovian regressors are not an issue, since the status at each time is entirely clear. Comparing this to the simulation approach, we can have a misprediction in one of two ways. 

\begin{enumerate}
\item The loan transitions to $ P $
\item The loan transitions to $ 3 $
\end{enumerate}

\paragraph{}

In the first case above, since $ P $ is an absorbing state, no harm has been done. Though the non-Markovian regressors are no longer correct, they are also no longer needed. In the second case, it would be necessary to then begin real simulation in order to correctly calculate future states. Taking $ T(t, X) $ to be the probability that the loan transitions from $ C $ to $ X $ in month $ t $, the following three values are needed. 

\begin{enumerate}
\item $P(t, C^*)$: The probability that the loan is always in $ C $
\item $P(t, P^*)$: The probability that the loan is in $ P $ after previously having always been $ C $
\item $T(t, 3^*)$: The probability that in the current month, the loan goes to $ 3 $ having always been $ C $
\end{enumerate}

\paragraph{}

These values can be easily represented in terms of the previous months' values. 

\begin{equation}
P(t, C^*) = P(t-1, C^*)T(t-1, C^*)
\end{equation}

Similar equations exist for $ P $ and for $ 3 $ as well. 

\paragraph{}

If the loan matures at time $ \tau $, then compute $ P(\tau, 3^*) $ (using formulas such as those above), which is the probability that the loan ever reaches state $ 3 $. For each time $ t $, compute 

\begin{equation}
T'(t, 3) = \frac{T(t, 3)}{P(\tau, 3)}
\end{equation}

\paragraph{}

This is the probability that the loan enters state $ 3 $ at time $ t $ given that it ever enters state $ 3 $. Now select a time $ t' $ randomly, weighted by $ T'(t, 3) $, and begin simulation from time $ t' $, given that the loan was current up until then. In this way, compute $ \vec{s'}(t) $ for all $ t \leq \tau $, taking $ \vec{s'}(t) = C $ where $ t < t' $. Similarly, define the status vector $ S(t) = \{P(t, C^*), P(t, P^*), 0\} $. Now the status vector $ \vec s(t) $ can be computed as 

\begin{equation}
\label{hybrid}
\vec s(t) = P(\tau, 3)\vec{s'}(t) + (1 - P(\tau, 3))S(t) 
\end{equation}

\paragraph{}

This computation has the advantages of both simulation and Markov matrix multiplication, and none of the weaknesses. On average, only approximately $ 1.5 $ columns of $ M $ need to be computed, rather than $ m $. In addition, non-Markovian regressors can be used freely, since in both branches of this computation, the complete state vector up to time $ t $ is always known with certainty. The simulation noise of this method is much, much less than the simulation noise introduced by raw simulation, since the rare events (transitions through $ 3 $) are over sampled, and the common events ($ C \rightarrow C$) are computed exactly. 

\paragraph{}

Lastly, this approach allows the practitioner to spend resources where they are most needed. I propose the following method, performed for each interest rate and house price path. For each loan number $ n $, compute $ w(n) = P(\tau, 3) $, with the understanding that $ P(\tau, 3) $ is a function of $ n $. 

\begin{enumerate}
\item Compute $ \gamma = \sum_{n} w(n) $, where the sum is taken over all $ n $ loans.
\item Determine some threshold, call it $ \varepsilon $, perhaps $ \varepsilon = \frac{\gamma}{nq} $ for some integer $ q $.
\item Loop through all loans, compute $ \alpha(n) = \min(1, \frac{w(n)}{\varepsilon}) $
\item For each loan, draw a random value $ \nu(n) \in [0,1) $
\item If $ \nu(n) > \alpha(n) $, skip this loan
\item Otherwise reduce $ w(n) $ by $ \varepsilon $, and compute a path (as defined in equation \ref{hybrid}) for this loan
\item Continue looping until all $ nq $ simulations have been performed, or no loan has $ w(n) > \frac{\varepsilon}{2} $. 
\end{enumerate}

\paragraph{}

This procedure is designed to allocate calculations efficiently. In general, loans that have a higher probability crossing status $ 3 $ will get more simulations. The end result is that all loans will have a similar amount of simulation noise at the end of the process. This avoids the very common problem where most loans have very little uncertainty, but a few have a lot, so most computational resources are wasted on the loans where it makes little difference. 

\paragraph{}

An alternate version would involve no randomness, and would simply apply $ \frac{w(n)}{\varepsilon} $ (rounded up, presumably) computations to each loan.

\section{Results}

\paragraph{}

The ITEM model was implemented as discussed above, and applied to the Freddie Mac loan level dataset. For the purposes of this demonstration, only a limited set of regressors were used. In a true production setting, a much larger set would be used, but the results would be otherwise very simliar. The inclusion of several of the most important regressors will be enough to prove the point.

\paragraph{}

Additionally, it should be noted that only the first few million observations from the database were used. The observation count is limited by the RAM of the test machine. In a live production setting, the observations used would be selected randomly, in addition, they would be shuffled. Without the shuffling, it is known that the observations used correspond to the oldest loans. In addition, the adaptive optimization is somewhat undermined by the fact that the first few blocks of calculations represent only a few (about 10k) loans originated around the same time. Therefore, no curve is reachable that doesn't improve the fit of these loans. This deficiency was allowed to remain in order to show the behavior of a multi-curve age regressor, which otherwise would have ended up with only two curves rather than the 5 shown here. 

\subsection{Initial Curve Fitting Convergence}

\paragraph{}

One important aspect of fitting is that it must be parsimonious with the curves to be added. The model performs well on this score. The table below describes the curves added when the model is allowed to fit on 1 million data points from the dataset. The model was allowed to select curves from age, incentive, fico, LTV and calendar month (to capture seasonal effects). Among these 5 regressors, 13 curves were drawn. The model was allowed to draw as many curves as it found useful, but stopped after 13. 

\paragraph{}

\begin{table}[H]
\caption {Curves (1M datapoints)}
\begin{center}
\begin{tabular}{|c|c|c|c|c|c|c|c|}
\hline
Regressor & toState & type & center & slope & negLL & AIC & BIC \\
\hline
age & P & logistic & 30.7 & 0.6 & 0.135644 & -7565.79 & -7530.34 \\
age & P & gaussian & 30.73 & 34.87 & 0.134044 & -2716.92 & -2681.47 \\
incentive & P & logistic & -0.18 & 1.01 & 0.132952 & -1775.78 & -1740.33 \\
age & 3 & logistic & 30.7 & 0.16 & 0.132498 & -632.64 & -597.19 \\
incentive & P & logistic & -0.18 & 0.99 & 0.132143 & -673.67 & -638.22 \\
... & ... & ... & ... & ... & ... & ... & ... \\
age & P & logistic & 30.72 & 0.265 & 0.130196 & -233.67 & -198.22 \\
... & ... & ... & ... & ... & ... & ... & ... \\
age & 3 & logistic & 30.72 & 0.074 & 0.130020 & -21.65 & 13.80 \\
\hline
\end{tabular}
\end{center}
\end{table}

\paragraph{}

\begin{figure}[H]
\centering
\includegraphics[scale=0.5]{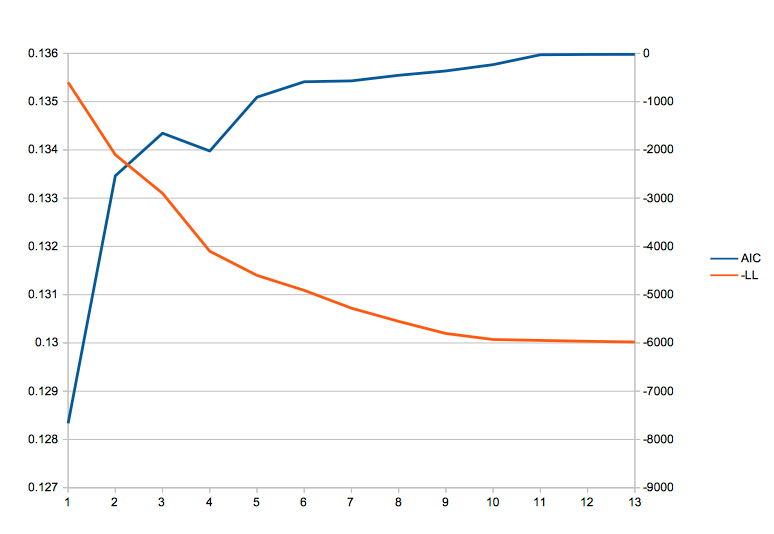}
    \caption{Fit Convergence as a Function of Curve Count}
    \label{fig:convergence}
\end{figure}

\paragraph{}

In the table above, the first 5 of these curves are shown, as well as curve 10 and curve 13. The AIC shows exponential decay behavior, though the decay slows for curves 6 through 10. It is interesting to note that curve 5 improves the AIC more than curve 4. The improvement in the age fit allowed for a better selection of incentive curve in the next step than was possible previously, this is a good example of the path dependency that annealing is designed to handle. After an intial improvement, the results quickly level off, as expected. 

\paragraph{}

The BIC for each of these curves is also recorded in the table. It differs from the AIC by a constant of roughly 35 for a dataset of this size. If the BIC is used instead of the AIC, then only 10 curves (rather than 13) would have been drawn. The difference in final log likelihood is immaterial ($ 0.130071 $ vs $ 0.130020 $), but the computational cost would have been reduced by roughly 20\%. 

\paragraph{}

After these 13 (or 10) curves, the model is unable to find any better curves. The primary reason for this is that the std. deviation of the log likelihood is large enough that all future optimizations rapidly halt, as there is not enough data to conclusively point the way to a better curve. With a larger dataset, it would be expected that this situation would improve. Even so, the curve count is expected to be highly sub-linear in dataset size. Notice also that ITEM did something that a typical human analyst would not do, it did not use all of the regressors, instead opting to place more curves on the most important few regressors rather than placing a few on on each.

\paragraph{}

The concentration of the curves on a few regressors is an interesting phenomenon. What is really happening here is that several of the regressors have strong colinearity and a dramatic difference in predictive power. Any weak regressor will appear to be pure noise if the distribution of stronger regressors is not uniform when projected onto this regressor. This prevents weaker regressors from getting any curves on them until the stronger regressors are adequately fit. At the same time, some strong regressors are strongly colinear with each other, meaning that each curve drawn on one of the regressors actually greatly improves the smoothness of the other regressor. Consequently, the algorithm moves back and forth between the two strong colinear regressors, ignoring all others until those regressors have been fully accounted for. You can see this oscillation in the order in which the curves were fit.

\paragraph{}

\begin{table}[H]
\caption {Curves (5M datapoints)}
\begin{center}
\begin{tabular}{|c|c|c|c|c|c|c|c|}
\hline
Regressor & toState & type & center & slope & negLL & AIC & BIC \\
\hline
age & P & logistic & 31.56 & 0.72 & 0.132417 & -38371.64 &-38331.37 \\
age & P & gaussian & 31.56 & 35.08 & 0.131104 & -11024.60 & -10984.33 \\
incentive & P & logistic & -0.04 & 1.92 & 0.129150 & -18802.34 & -18762.07 \\
age & P & gaussisn & 31.56 & 35.08 & 0.128644 & -4714.76 & -4674.49 \\
age & P & logistic & 31.57 & 0.89 & 0.128296 & -3300.35 & -3260.08 \\
age & 3 & logistic & 31.57 & 0.13 & 0.128080 & -2074.88 & -2034.61 \\
ltv & 3 & logistic & 74.05 & 0.55 & 0.127905 & -1676.31 & -1636.04 \\
incentive & 3 & logistic & -0.14 & 1.09 & 0.127570 & -1994.14 & -1953.87 \\
age & P & logistic & 31.57 & 0.06 & 0.127537 & -241.15 & -200.88 \\
\hline
\end{tabular}
\end{center}
\end{table}

\paragraph{}

In a larger dataset, with 5 million points (see table above), fewer curves are added, but better results are achieved. The reason for this is related to the defense against overfitting built into the model. The model will not refine a parameter beyond the point where it moves the log likelihood by less than one sample standard deviation. This protects against over fitting by avoiding cures that might otherwise have a spuriously high AIC score. In a larger dataset, not only is the noise within the dataset suppressed (thus allowing easier fitting to the smooth model curves), but the parameters can be refined further due to the smaller standard deviation of the mean of the log likelihood. This means that each curve is in general a better fit, which can offset the ability of the dataset to otherwise support additional curves. Also, note that there was no difference between AIC and BIC, they both would have included the same set of curves. 

\paragraph{}

Note here that the centrality parameter of these curves shows little variation. This is apparently due to the presence of numerous local minima in the noisy dataset. One solution to this problem is to pick the centrality parameter by simply selecting the relevant value from a random point in the dataset. This will have the effect of starting the curves (typically) in concentrated parts of the dataset, but the added noise should help to overcome this problem related to local minima.

\subsection{Fitting Computational Cost}

For the purposes of this test, the fitting is allowed to draw any number of curves on any of 5 regressors for 3 transitions. This fitting was performed several times with different dataset sizes to see how the fitting time grows with dataset size. It was also retried with several configuration options related to adaptive optimization to measure the performance with and without the improvements brought about by adaptive optimization. All tests were performed on a 2009 Mac Pro with 4 physical CPU cores, 8 virtual cores. 

\paragraph{}

\begin{table}[H]
\caption {Fitting Computational Cost}
\begin{center}
\begin{tabular}{|c|c|c|c|c|c|c|c|}
\hline
Adaptive & Sigma Stop & Observations & Curves & -LL & Time/Curve (ms) & Time (ms) & Curve 8 (ms) \\
\hline
TRUE & TRUE & $ 1 * 10^{6} $ & 13 & 0.1300 & 165,000 & 2,147,031 & 855,181 \\
TRUE & TRUE & $ 5 * 10^{6} $ & 9 & 0.1275 & 514,000 & 4,626,543 & 4,103,329 \\
TRUE & TRUE & $ 1 * 10^{7} $ & 8 & 0.1284 & 983,000 & 7,864,467 & 7,864,467 \\
FALSE & TRUE & $ 1 * 10^{6} $ & 19 & 0.1299 & 115,707 & 2,198,424 & 1,008,361 \\
FALSE & TRUE & $ 5 * 10^{6} $ & 16 & 0.1272 & 370,584 & 7,041,099 & 3,499,648 \\
FALSE & TRUE & $ 1 * 10^{7} $ & 18 & 0.1280 & 910,426 & 17,298,097 & 8,283,654 \\
FALSE & FALSE & $ 1 * 10^{6} $ & 16 & 0.1299 & 112,765 & 2,142,544 & 1,008,361 \\
FALSE & FALSE & $ 5 * 10^{6} $ & 16 & 0.1272 & 592,825 & 11,263,666 & 5,652,720 \\
FALSE & FALSE & $ 1 * 10^{7} $ & 18 & 0.1280 & 1,271,021 & 24,149,392 & 10,042,209 \\
\hline
\end{tabular}
\end{center}
\end{table}

\paragraph{}

The column Adaptive notes whether or not this test optimizes by examining the whole dataset on each iteration, or simply examines a subset if that is enough to determine which point is better. The column "Sigma Stop" indicates whether or not the optimization stops when the points are all within one standard deviation of each other.

\paragraph{}

The table above includes three time estimates. One is the time per curve, another is the total time, and the third is the time required to draw 8 curves. This last (curve 8) time is perhaps the fairest measure. The values are noisy, so it is hard to draw any firm conclusions, but the adaptive optimization does generally perform better. The total time is much better for the adaptive optimization, since it avoids drawing a large number of relatively less important curves for larger dataset sizes.  Notice that the rows where both of these flags are false indicate typical optimization algorithms. The adaptive optimization with a 1-sigma stop performs almost 3x faster (20\% faster on a curve for curve basis) already by the time the dataset gets to 10 million points. With a much larger dataset near a billion points, this difference would become much, much larger. The adaptive optimization does not lose any meaningful degree of accuracy, and in fact is far more parsimonious with curves as well simply due to the fact that it avoids fitting curves that cannot match well on several subsets of the data set. 

\paragraph{}

It is expected that this trend continues into very large dataset sizes, saving substantial computation time. In this case, ten times more data requires approximately 4 times as much computation time. For very large dataset sizes, the cost of adaptive optimization (at a given level of precision) would approach a constant. It is expected that the cost of a round of annealing would be similar to the cost of drawing all these curves initially. 

\paragraph{}

The real limitation in this context is RAM. The calculation requires approximately 500 MB of RAM per million observations. Given that the computer used for this exercise has only 8 GB of RAM, it is hard to push much beyond 10 million observations, or roughly 1\% of the dataset. Similarly, for reasons of simplicity, this sample is not being chosen uniformly from the entire dataset, but is rather composed of the first observations from the set. This is done just for speed and efficiency when loading the data. In a real production setting, a representative sample (possibly the whole dataset) would be precompiled, and that would then be used. It is estimated that a computer with roughly 500-1000 GB of RAM (and about 40 CPU cores) could optimize a model over the entire dataset. A system such as this is well within the reach of even a small corporation, and might cost approximately \$50,000. Such an optimization would take roughly 6 hours, easily accomplished as an overnight job. Smaller optimizations on representative subsets of the data could be finished in minutes, allowing for very rapid prototyping and development. 

\paragraph{}

\begin{table}[H]
\caption {Cost Per Row}
\begin{center}
\begin{tabular}{|c|c|c|c|c|c|}
\hline
Fit Type & Accuracy & Observations & rows/ms & ns/row & cycles/row \\
\hline
Curve & Correctly Rounded & $ 10^{6} $ & $ 6042 $ & $ 165.5 $ & $ 1760 $ \\
Curve & Approx $ 10^{-15} $ & $ 10^{6} $ & $ 18416 $ & $ 54.3 $ & $ 577 $ \\
Coefficient & Correctly Rounded & $ 10^{6} $ & $ 1095 $ & $ 9127 $ & $ 9711 $ \\
Coefficient & Approx $ 10^{-15} $ & $ 10^{6} $ & $ 1469 $ & $ 6803 $ & $ 7238 $ \\
\hline
\end{tabular}
\end{center}
\end{table}

\paragraph{}

As can be seen from the table above, the cost per row of these optimizations is very moderate. It requires roughly 1700 clock cycles to do a row of curve fitting, and roughly 10,000 to do a row of coefficient optimization. In this case, the coefficient optimization was fitting only flags, that cost would be expected to rise somewhat as more curves are introduced. The curve fitting cost is far more important, as more than 90\% of the fitting time is consumed by curve fitting rather than coefficient fitting. 

\paragraph{}

The code used for these examples is not fully optimized. In particular, it uses correctly rounded $ e^{x} $ and $ \ln $ functions. Replacing the correctly rounded exponential function with one with a relative error of approximately $ 10^{-15} $ more than triples the performance of the computation at no material cost to accuracy. Additional optimizations are certainly possible. The cycles/row computation assumes 100\% utilization of all four cores in the CPU of the test machine, but in practice the utilization rate is somewhat lower (about 70\%), primarily due to the small size of the dataset preventing a more efficient division. Additional tuning would increase that value to nearly 100\%, reducing the cost still further. For the curve fitting, it should be possible to process a single observation in less than 200 clock cycles on a typical modern CPU in Java. Straight C or C++ code would perform similarly, though it might be possible to do better with carefully hand tuned assembly code that takes full advantage of the SIMD units within newer CPUs. 

\paragraph{}

Similarly, the coefficient optimization is essentially the same operation that would be used to project future behavior. Even with no additional code improvements, this cost is in line with the resource budget set out at the beginning of the paper. Careful code optimization should keep the calculations under budget even in a full scale model using numerous regressors and interaction terms.

\subsection{Fitting Results}
\label{fitResult}

With the example above (using 1 million data points), already the results are greatly improved by the curves that were drawn. Notice that all of these fits were computed entirely without human intervention other than for a human to define the regressors, and give the model the list of regressors it may use. A typical mortgage model built within industry takes at least a man-month of data analysis, fiting and validation. The ITEM model automates most of these tasks, accomplishing in minutes what a human would typically do in a few days in a more traditional setting.

\paragraph{}

\begin{figure}[H]
\centering
\includegraphics[scale=0.5]{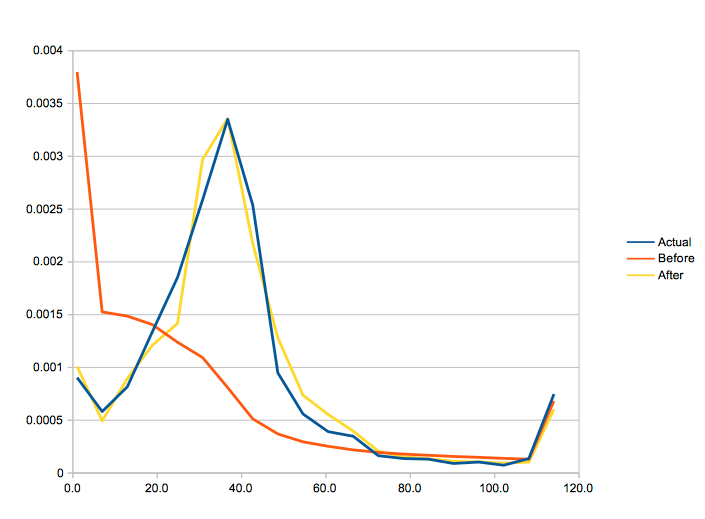}
    \caption{C to P vs. Age}
    \label{fig:ageBeforeAfter}
\end{figure}

\begin{figure}[H]
\centering
\includegraphics[scale=0.5]{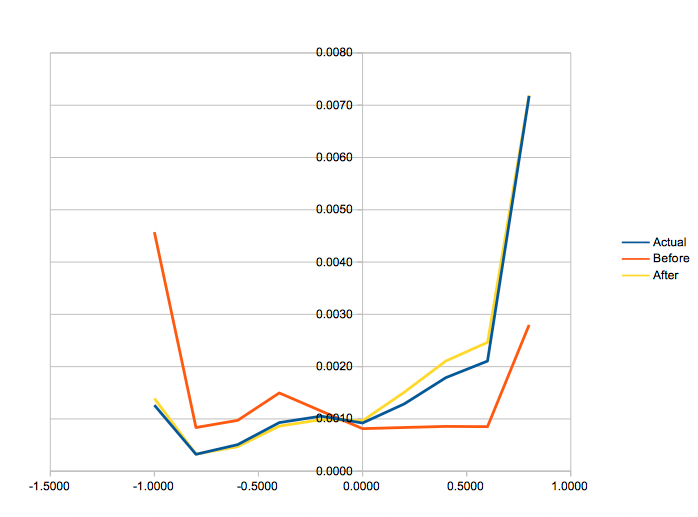}
    \caption{C to P vs. Incentive}
    \label{fig:incentBeforeAfter}
\end{figure}

\paragraph{}

In these charts, the curve labled Before is from the model with all flags fit, but before any of the curves have been drawn. So this model does not include the effect of age or incentive. The curve labeled After is from the model with the curves added. The charts show clearly that even a few curves have dramatically improved the fitting of these regressors. 

\paragraph{}

Note that the age curve shown is strongly influenced by the fact that the loans under consideration are all originated near the same time. Therefore, the spike seen around age 36 months is strongly influenced by the prevailing interest rates at that time. With a larger dataset, this correlation between age and time would smooth out, and a more classic age curve would be recovered. However, for the purposes of this demonstration, it is enough to show that ITEM can construct a model that represents the data. It is not necessary to show that the data set chosen is a truly fair representation of the mortgage universe. 

\paragraph{}

The seasonality curve shows a good example where again annealing might be helpful. 

\paragraph{}

\begin{figure}[H]
\centering
\includegraphics[scale=0.5]{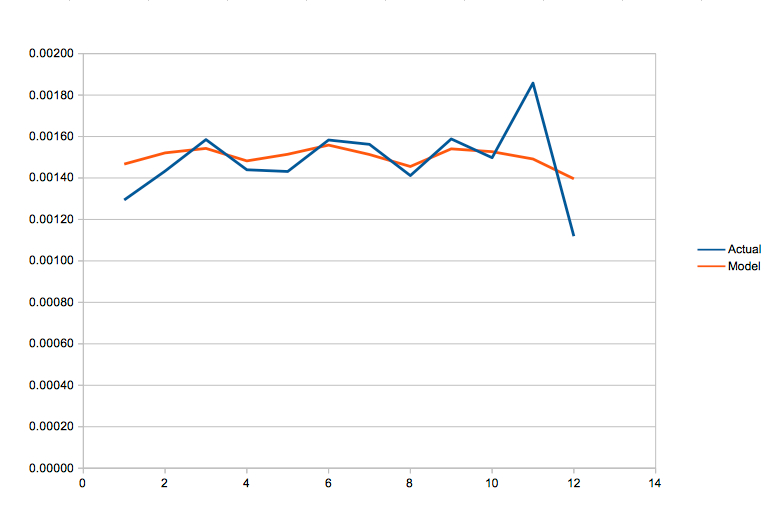}
    \caption{C to P vs. Calendar Month}
    \label{fig:seasonBeforeAfter}
\end{figure}

\paragraph{}

It can be seen in this example that the seasonality effect was not captured. Ideally, that spike near the end should be captured, but give all the smaller spikes throughout the year, the optimizer would be likely to hit a very unsatisfying local minimum rather than finding the actual behavior related to December and January. Also, in this example, the low season for prepayment spans the year end, so it is not localized when looking at the seasonality in this way. The model might still draw curves here that would help, but it would take a large amount of data and would take at least two curves to really improve this significantly. This is an example where a small amount of randomness could help dramatically, by allowing the optimizer to find the spike at the end. 

\subsection{Model Curves}

A full and complete examination of all curves drawn by this model is beyond the scope of this paper. However, for illustrative purposes, the curve related to incentive for the 5 million point dataset is shown below. 

\paragraph{}

\begin{figure}[H]
\centering
\includegraphics[scale=0.5]{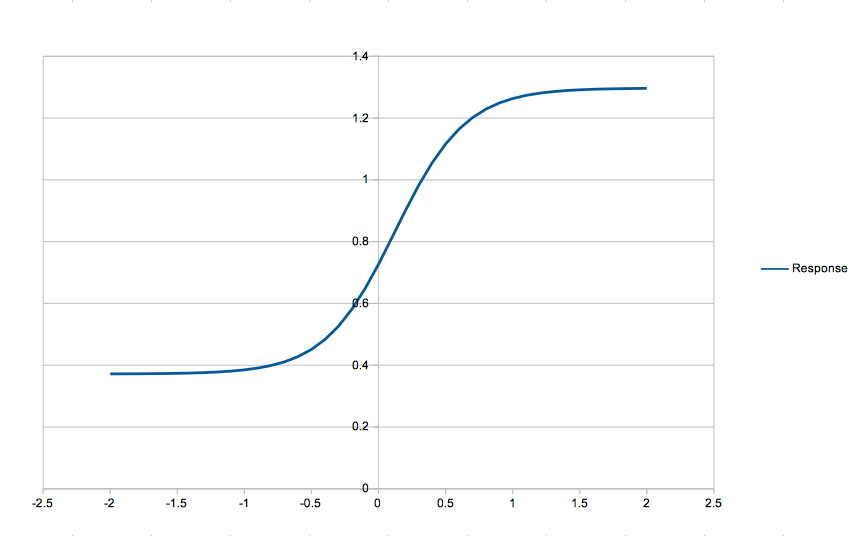}
    \caption{C to P multiple vs. Incentive}
    \label{fig:incentResponse}
\end{figure}

\paragraph{}

As can be seen from Figure \ref{fig:incentResponse}, the response of the model to incentive is an extremely smooth curve that rapidly saturates in both the positive and negative directions. This curve has been represented here as a multiple. Assuming the $ C -> P $ probability is not too large, then with a given value of incentive, that probability is multiplied by the value of this curve. Only relative multiples matter here, so it doesn't matter if the curve passes through 1 or not. This shows that a loan with strong negative incentive is about one third as likely to prepay as a loan with strong positive incentive. This fits with intuition, and in fact is a close approximation of the historical data seen in Figure \ref{fig:incentBeforeAfter}. Notice that in Figure \ref{fig:incentBeforeAfter} the highest vs. lowest prepayment multiple is approximately 7 to 1, but that graph is confounded by numerous other factors (occupancy type, age, etc...), whereas the response above is purely for incentive acting on its own. 

\paragraph{}

The primary advantage of the ITEM model is that it expresses all responses to input variables in terms of these extremely smooth curves, and thus tends not to over fit. This can be seen in the response to Age, which has 5 different curves fit to it for the $ C -> P $ transition. 

\paragraph{}

\begin{figure}[H]
\centering
\includegraphics[scale=0.5]{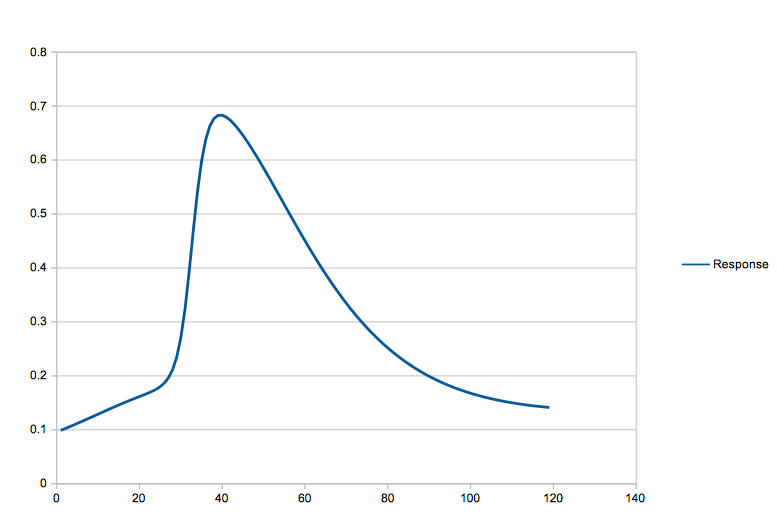}
    \caption{C to P multiple vs. Age}
    \label{fig:ageResponse}
\end{figure}

\paragraph{}

Even here, the curves combine to make a single very smooth curve. This response is largely driven by the correlation between age and interest rates, due to the limited fitting sample. Arguably, this is a worst case scenario. As the sample size increases, and more variation in the loan-months is pulled in, this age curve will flatten out and look smoother. However, even as it is, the age curve makes intuitive sense. Loans tend to not prepay for the first few years of life, then they enter a period of fast prepays before finally tailing off into old age. This curve captures that phenomenon surprisingly well already. 

\section{Advanced Fitting Topics}

\paragraph{}

The previous sections explored some basic fitting routines and results. This section will concentrate on a single large fit (roughly 15 million data points) drawn randomly from the dataset. The loans in this dataset are selected by hasing the loan id from the dataset using SHA-256, then reducing it mod 50 and taking any loan with a reduction congruent to 0. This selection procedure should ensure a well distributed random sample. 

 \paragraph{}

In this exercise, several experimental fitting choices were made in an attempt to improve the fit. 

\begin{enumerate}
\item The ordering of the data points was randomized
\item The starting curve centrality parameter was set to the regressor value of a random selection from the sample
\item The starting curve slope parameter was randomized in sign and magnitude
\item Annealing was run at the conclusion of the initial round of fitting
\end{enumerate}

 \paragraph{}

The addition of these random choices would be expected to make the model less likely to get caught up in local minima, and indeed the quality of the fit is qualitatively better than those seen before. One primary reason for this is that the optimizer finds a saddle point at zero beta, which means that it has a very difficult time reversing the sign of beta during the optimization. Always starting the beta as a positive number would therefore tend to bias the selection strongly towards results that would naturally have a positive beta, reducing the fit quality. Using a random starting point and random sign of beta helps this problem greatly. 

\paragraph{}

One unexpected side-effect is that gaussian curves become significantly more common than logistic curves. The likely reason for this is that since the beta sign cannot be easily flipped, a logistic that starts off with the wrong sign will be unable to improve the results. However, a Gaussian can overcome the incorrect beta sign by simply moving the centrality parameter to be very high or very low. In the future, perhaps an adjustment to attempt both signs on every iteration would reduce this bias towards Gaussians. 

\paragraph{}

Surprisingly, annealing was unable to improve the quality of this fit. The issue appears to be the high degree of colinearity in the regressors. With so much colinearity, dropping all the curves from a given regressor does not provide a very clean slate, and it may be found that only curves that interlock with existing curves in a very particular way are very effective. Presumably, annealing would be more effective if it was used earlier in the process, as here it was used only at the end. In addition, in a less contrived example where more regressors are considered, it is expected that annealing will prove more useful. 

 \paragraph{}

The tables below show the in sample validation against several key variables. Notice that the qualitative fit is not much better, but the regressors (especially age) that are strongly correlated with time make a lot more sense due to the better sampling that reduces the contemporality of the selected loans. 

\begin{figure}[H]
\centering
\includegraphics[scale=0.5]{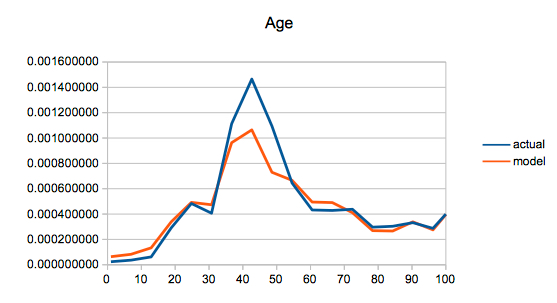}
    \caption{C to P vs. Age}
    \label{fig:ageValidation}
\end{figure}

 \paragraph{}

\begin{figure}[H]
\centering
\includegraphics[scale=0.5]{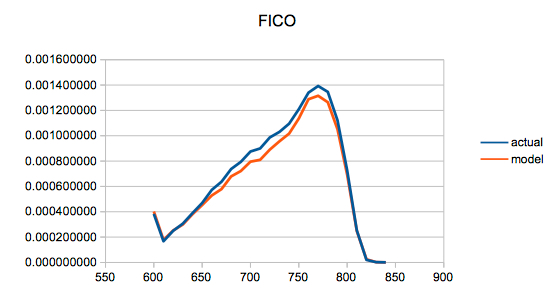}
    \caption{C to P vs. FICO}
    \label{fig:ficoValidation}
\end{figure}

 \paragraph{}

\begin{figure}[H]
\centering
\includegraphics[scale=0.5]{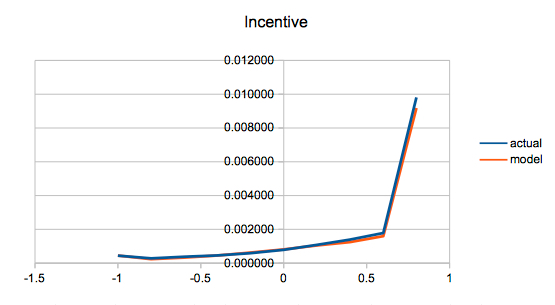}
    \caption{C to P vs. Incentive}
    \label{fig:incentValidation}
\end{figure}

 \paragraph{}

\begin{figure}[H]
\centering
\includegraphics[scale=0.5]{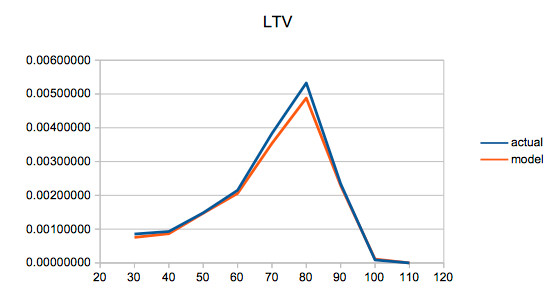}
    \caption{C to P vs. LTV}
    \label{fig:ltvValidation}
\end{figure}

 \paragraph{}

The one curve that stands out is LTV. Though the model was allowed to draw curves on LTV, it did not do so, yet the LTV results fit very well. Primarily, this is due to colinearity between LTV and other regressors, primarily FICO. If more up-to-date LTV values were available, this regressor woudl be far more useful. Below, the model curves for these regressors are shown. Here the model transition probabilities are charted directly for some sensible choice of the regressors not being charted. The effect of the excessive use of Gaussians is immediately obvious, as the curves start to do unusual things once they leave the domain where the data lies. For loans that could be extreme outliers, this could matter. A solution as simple as capping/flooring each regressor at the 1-percentile level would suffice to eliminate any danger here. 

 \paragraph{}

\begin{figure}[H]
\centering
\includegraphics[scale=0.5]{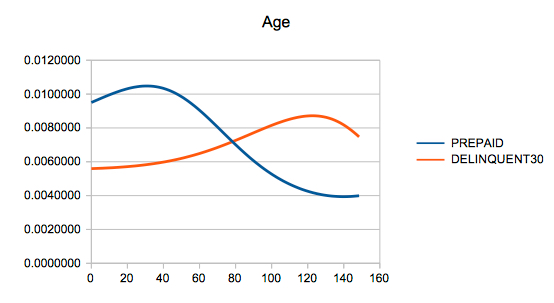}
    \caption{Transition Probability by Age}
    \label{fig:ageCurve}
\end{figure}

 \paragraph{}

\begin{figure}[H]
\centering
\includegraphics[scale=0.5]{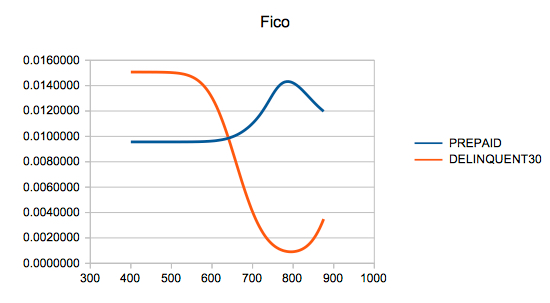}
    \caption{Transition Probability by Fico}
    \label{fig:ficoCurve}
\end{figure}

 \paragraph{}

\begin{figure}[H]
\centering
\includegraphics[scale=0.5]{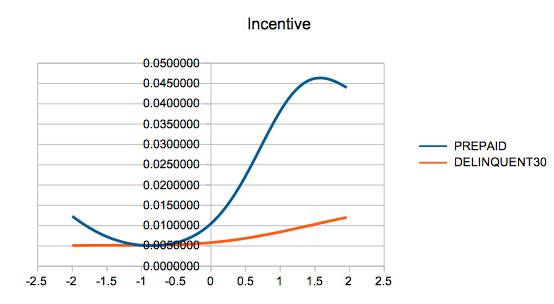}
    \caption{Transition Probability by Incentive}
    \label{fig:incentCurve}
\end{figure}

 \paragraph{}

\begin{figure}[H]
\centering
\includegraphics[scale=0.5]{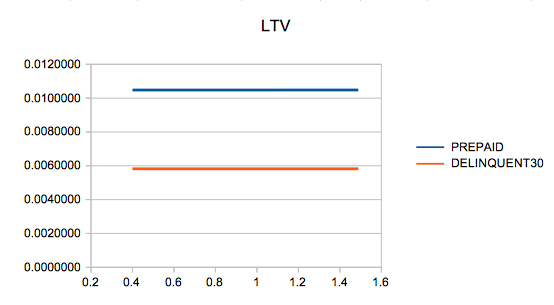}
    \caption{Transition Probability by LTV}
    \label{fig:ltvCurve}
\end{figure}

\paragraph{}

Notice in particular how FICO begins to bend back after roughly 800. There is simply no appreciable level of defaults at that FICO level, and very little data of any form. Improvements to the fitting routines might eliminate the usage of Gaussians there, and greatly reduce the propensity for this sort of issue. A similar story holds for incentive. 

\paragraph{}

Again, the important result here is not that these curves are dramatically better than other curves show by other mortgage researchers. The advantage of this approach is that the curves are generated with no human intervention. Using this as the starting point, it is easy to correct minor issues (e.g. cap and floor some regressors) and thus very quickly arrive at a very high quality model.

\section{Conclusion}

The ITEM model was designed to automate repetitive tasks and allow for an efficient model to be built automatically. In the examples above, all fits were performed automatically, with no manual intervention beyond the initial definition of the regressors. In a more typical mortgage model, a modeler would typically spend months analyzing data, selecting regressors, and examining model residuals. Each iteration of this process would require several days, and many dead ends would be explored, largely due to inefficient regressor and parameter choices. Generally, another process would be required to remove insignificant parameters, which would then be followed up again by further rounds of fitting and analysis. 

\paragraph{}

The ITEM model automates this entire cycle. It runs thousands of iterations of fit, extend, analyze residuals, and then fit again, all within the space of minutes rather than days. The model does not add insignificant parameters, so there are none to be removed, though some annealing may be needed. Similarly, ITEM does not choose suboptimal curves or regressors, so it explores fewer dead ends. Since the model is additive in logit space, projecting it onto any individual regressor gives a fair representation of its behavior, greatly easing the job of validation. 

\paragraph{}

The workflowing using ITEM is to simply select the dataset, define the regressors, allow ITEM to fit a model, and then go straight to final model validation. 

\section{Reference Implementation}

\paragraph{}

Attached to this submission is a Java reference implementation of the ITEM model core. This reference implementation provides only the core modeling functionality. In order to run this model, the practitioner will need to define enumerations (see edu.columbia.tjw.item.base.StandardCurveType as an example) describing the statuses available to the modeled phenomenon, and also describing the regressors available. In addition, the practitioner will need to provide a class to produce grids of data implementing edu.columbia.tjw.item.ItemFittingGrid, and related interfaces.

\paragraph{}

Once these interfaces are implemented, the model may be fit using the ItemFitter. The resulting model is suitable for projection, but no general projection code is provided in this reference implementation. Similarly, at this time, the annealing code is not provided, but could be made available if there is interest. 

\paragraph{}

This reference implementation is made available as a Maven package:

\begin{itemize}
\item groupId: edu.columbia.tjw
\item artifactId: item
\item version: 1.0
\end{itemize}

\paragraph{}

If there is interest, this archive could be published to the Maven central repository.

\section{Future Work}

Anticipated future work includes some results for annealing passes and the effect of noise introduction during fitting. In addition, the model could be applied to a larger set of regressors to show a more fully featured mortgage model. Applying the model to large regressor sets can reveal previously unknown behavior which may be worth research in its own right. The hybrid simulation approach could also be investigated more fully.

\pagebreak

\addcontentsline{toc}{section}{References}

\bibliographystyle{acm}

\pagebreak

\begin{appendix}

\renewcommand{\thesection}{\Roman{section}}

\section{Appendix: Information-Theoretic Efficiency}
\renewcommand{\thesubsection}{\Alph{subsection}}

\paragraph{}

This paper mentions information-theoretic efficiency in several places. This term can have several meaning, so this appendix will define the meaning for the purposes of this paper. 

\paragraph{}

The first thing to note about modeling in general, is that it is closely related to the concept of relative entropy, i.e. entropy against a party that has access to some models or data. In fact, a model is little more than a compression function, with all the limitations that entails. For instance, suppose the datasets $ X $ and $ Y $ are available. If one wished to transmit this data to another person, it might be more efficient to first fit a model $ A_{XY}(\vec x) = \vec y + \varepsilon $, and then transmit $ X $, $ A_{XY} $, and $ \varepsilon $. Ideally, the entropy of the model and the residual together would be smaller than the entropy of the response $ \vec y $. We know, however, that any compression algorithm that makes at least one dataset shorter must make at least one dataset longer. So it is known apriori that there are some datasets for which this model will give such bad predictions that it actually increases the entropy. Therefore, there can be no perfect model, as any model could be defeated by carefully chosen data. 

\paragraph{}

Because of this, most of the approaches in this paper are heuristic, with some motivation from fields such as analysis but no complete proofs. 

\paragraph{}

The relative entropy of a data set is simply its log likelihood (up to some constant factors), so a model that improves the log likelihood fo the dataset enough, would be worthwhile, and one that doesn't is not. Various information criteria (e.g. AIC, BIC) are based on a calculation of the total entropy contained in the parameter set $ \theta $ and the residuals $ \varepsilon $, and comparing that value to what it would be for an alternate parameters set $ \theta' $ and $ \varepsilon' $. The best model is then the one with the lowest AIC, it has extracted as much information as possible from the dataset without passing the point where additional model complexity is counter productive. 

\paragraph{}

With this in mind, suppose that in the typical fashion, the phenomenon actually follows some well defined but unknown distribution, with some associated noise term. Assume that the noise is pure noise, in that no model can improve its entropy, taking into consideration the entropy of the model itself. 

\begin{equation}
\vec y = f(\vec x, \vec w) + \varepsilon
\end{equation}

\paragraph{}

Here, $ \vec x $ is some observable data, but $ \vec w $ is unobservable. This situation might be simplified by absorbing the unobservable information into the noise term $ \varepsilon $. The noise terms may no longer be i.i.d., but they should at least be mean zero.

\begin{equation}
\vec y = f(\vec x) + \varepsilon
\end{equation}

\paragraph{}

N.B. that the noise term is no longer pure noise, it is possible that a model could reduce it if there is correlation between $ \vec w $ and $ \vec x $. This will result in a lower log likelihood, but will also result in a misattribution of some behavior away from $ \vec w $ towards $ \vec x $. However, being aware of the dangers, suppose the model assumes that $ \vec y $ follows some parameterized distribution $ g $ with some unknown parameters $ \theta $. 

\begin{equation}
\vec y \approx g(\vec x | \theta) + \varepsilon
\end{equation}

\paragraph{}

The parameter (in practice, perhaps many parameters)  $ \theta $ is unknown, but it could be estimated, perhaps with maximum likelihood estimation. This would recover an estimator for $ \theta $, namely $ \hat \theta $. Now, to state that a model is information theoretically efficient, we will in this paper mean the combination of three requirements, expressed in terms of the above quantities and also the sample size $ N $. 

\begin{enumerate}
\item $ \lim_{N \rightarrow \infty} g(\vec x | \theta) = f(\vec x) $
\item $ \lim_{N \rightarrow \infty} \hat \theta = \theta $
\item The variance of $ \hat \theta $ reaches the Cramer-Rao bound asymptotically.
\end{enumerate}

\paragraph{}

Briefly, requirements (2) and (3) above are the standard requirements for an efficient estimator. The first condition states simply that the model $ g $ will converge to the actual function defining the phenomenon, given enough data, i.e. the model is not misspecified. For parameterized models, this is almost never true. Only in extremely rare circumstances would it be the case that the actual physical phenomenon being modeled converges to exactly the functional form chosen for the model. As the dataset becomes large, the majority of the residual error in the model would be due to this misspecification.  

\paragraph{}

If the model is known to be misspecified and does not converge to the true distribution, then it may still be possible to define a sequence of models such that. 

\begin{equation}
\lim_{n \rightarrow \infty} g_{n}(\vec x | \theta_{n}) = f(\vec x)
\end{equation}

\paragraph{}

Where here, the varable $ n $ is determining how many variables are included in the model. In other words, $ g_{n}(\vec x | \theta_{n}) $ is now a family of models with an unbounded number of parameters. In this case, information theoretic efficiency is expanded to include now four separate items. It is understood that all the following equations hold only as $ N \rightarrow \infty $.

\begin{enumerate}
\item $ \lim_{n \rightarrow \infty} g_{n}(\vec x | \theta_{n}) = f(\vec x) $
\item $ | g_{n}(\vec x | \theta_{n}) - f(\vec x) | $ is minimized for all $ n $
\item $ \hat \theta = \theta $
\item The variance of $ \hat \theta $ reaches the Cramer-Rao bound asymptotically.
\end{enumerate}

\paragraph{}

Where again, the first item is ideal but unlikely in practice. Similarly, the second item is an idealization, but can be taken to mean just that the model is parsimonious with parameters, getting meaningful improvement with each parameter added. The last two items are as before, measures of unbiased and asymptotically efficient estimation. A maximum likelihood estimation is (under most conditions) guaranteed to be both asymptotically efficient and unbiased, so the last two conditions will then be satisfied. Any parameterized model is also virtually guaranteed to not converge to exactly the actual phenomenon under study. What remains then is to consider the distance between the model and the phenomenon for any given number of parameters, and when information theoretic efficiency is discussed, this is typically what will be meant. For this notion of closeness, a typical metric would be Kullback-Leibler divergence though any equivalent metric would do.  

\paragraph{}

Since no model can be perfect due to the compression arguments above, there is no true solution to this problem and no ideal model. It is therefore necessary to from here proceed with heuristics for which there is some theoretical basis to believe that they may in many cases produce good models. One common assumption is that $ f(\vec x) $ is smooth and continuous with respect to $ \vec x $, depending on the model family, other assumptions may be needed as well.

\end{appendix}

\end{document}